\documentclass[10pt, twocolumn, journal]{IEEEtran}

\usepackage{cite}
\usepackage{graphicx}
\usepackage{epsf}
\usepackage{subfigure}
\usepackage{amsmath}
\usepackage{comment}
\usepackage{amssymb}
\usepackage{array}
\usepackage{setspace}
\usepackage{amsthm,amssymb}
\usepackage[ruled,lined]{algorithm2e}
\usepackage{multirow}
\usepackage{tabularx}
\usepackage{xfrac}
\usepackage{breqn}
\usepackage{bbm}
\usepackage{enumerate}
\usepackage{algorithmic}
\usepackage{dsfont}
\usepackage{diagbox}
\usepackage{mathtools}
\usepackage{adjustbox}
\usepackage{color}

\graphicspath{{Fig/},{fig/}}

\newcommand{\fig}{Fig.\xspace}
\newcommand{\netw}{TCD-FERN\xspace}
\newcommand{\preproc}{DaS\xspace}

\title{\Huge Time-Selective RNN for Device-Free Multi-Room Human Presence Detection Using WiFi CSI} 

\author{ 
Li-Hsiang Shen,~\IEEEmembership{Member,~IEEE}, An-Hung Hsiao, Fang-Yu Chu, and Kai-Ten Feng,~\IEEEmembership{Senior Member,~IEEE}}

\begin{document}

\maketitle

\begin{abstract}
Device-free human presence detection is a crucial technology for various applications, including home automation, security, and healthcare. While camera-based systems have traditionally been used for this purpose, they raise privacy concerns. To address this issue, recent research has explored the use of wireless channel state information (CSI) extracted from commercial WiFi access points (APs) to provide detailed channel characteristics. In this paper, we propose a device-free human presence detection system for multi-room scenarios using a time-selective conditional dual feature extract recurrent network (TCD-FERN). Our system is designed to capture significant time features on current human features using a dynamic and static data preprocessing technique. We extract both moving and spatial features of people and differentiate between line-of-sight (LoS) and non-line-of-sight (NLoS) cases. Subcarrier fusion is carried out in order to provide more objective variation of each sample while reducing the computational complexity. A voting scheme is further adopted to mitigate the feature attenuation problem caused by room partitions, with around $3\%$ improvement of human presence detection accuracy. Experimental results have revealed the significant improvement of leveraging subcarrier fusion, dual-feature recurrent network, time selection and condition mechanisms. Compared to the existing works in open literature, our proposed TCD-FERN system can achieve above $97\%$ of human presence detection accuracy for multi-room scenarios with the adoption of fewer WiFi APs.
\end{abstract}


{\let\thefootnote\relax\footnotetext
{Li-Hsiang Shen is with Department of Communication Engineering, National Central University, Taoyuan, Taiwan (email: gp3xu4vu6@gmail.com)}}

{\let\thefootnote\relax\footnotetext
{An-Hung Hsiao, Fang-Yu Chu and Kai-Ten Feng are with the Department of Electronics and Electrical Engineering, National Yang Ming Chiao Tung University (NYCU), Hsinchu, Taiwan. (email: e.c@nycu.edu.tw, franceschuu@gmail.com, and ktfeng@nycu.edu.tw)}}

\section{Introduction}

Smart homes have become increasingly popular due to the pursuit of convenience and innovation. Home security is a crucial issue that requires attention, including intruder alerts, anti-theft, and crisis alarms. To ensure home safety, the stability and precision of the monitoring system are indispensable. Traditionally, wearable sensors are popular for positioning due to their high accuracy. However, these approaches require users to attach sensors to themselves \cite{ble_sense, phone_sen}, which can be inconvenient and bothersome in some situations. Moreover, wearable devices may require frequent battery recharging \cite{HanzoLifeTime}. Instead of using device-based methods, research on device-free \cite{DL_detect} schemes has increased, such as camera-based detection, infrared or radar sensors. Camera-based systems \cite{31, iotj_camera} provide particular information about human actions, height, or even facial characteristics, but they have privacy defects and blind spot problems. Infrared and radar sensor approaches overcome the privacy issue of camera-based systems \cite{radar_fall}, but they require humans to walk in a line-of-sight (LoS) path, which cannot detect their presence when staying in a non-line-of-sight (NLoS) path. Therefore, enhancing signal processing techniques in the physical layer is necessary to improve the attainable device-free detection precision at reduced computational complexity \cite{ourcsi1, ourcsi2, wifi_fall}.

Moreover, radio frequency (RF)-based signals are widely used due to availability from commodity WiFi devices without the need for dedicated devices \cite{wifi_loc, DL_detect, wifi_sense2}. This solves the issues of invasion and NLoS weakness \cite{acm} and results in convenience. The most common signal used is the received signal strength indicator (RSSI), which measures the received signal power. Although RSSI-based detection is useful in indoor localization \cite{1,2,finger}, human activity recognition \cite{3}, and gesture recognition \cite{4}, it only provides coarse-grained performance and can be affected by the multipath effect. To address these limitations, researchers have developed a fine-grained channel measurement technique called channel state information (CSI) \cite{5, DL_detect, acm, wifi_sense2}, which provides both time domain and frequency domain information. CSI has been used for various applications such as indoor localization \cite{6}, fall detection \cite{7,8}, human activity recognition \cite{9,10}, gesture detection \cite{11,12}, and respiration tracking \cite{13,14}. However, carrying out multi-object tracking using CSI is challenging due to the complexity of overlapped signals in temporal and spectral domains. Additionally, environmental monitoring techniques such as crowd counting \cite{15} and intrusion detection \cite{16} are popular with CSI-based methods, but they require a room to be equipped with both a transmitter (TX) and a receiver (RX), which is costly and impractical.

Presence detection, which infers the presence of people in indoor environments based on the variation of received signals \cite{detect2, detect2-1}, faces the challenge of false detection caused by a range of events that result in time-varying signals. For instance, humidity and temperature changes in the air, as well as the unpredictable locations of interfering objects, can cause false detection, necessitating time-consuming replenishment of the database. Previous studies have deployed a pair of TX/RX WiFi devices for presence detection in a single space in a small area \cite{sense_bm1} or have addressed the fine-grained signal processing for presence detection in a single conference room \cite{mycolor}. However, presence detection across different rooms \cite{ourcsichu, ourcsiKI} poses substantial challenges on the associated signal analysis, where both the signal attenuation and multipath effects need to be jointly taken into consideration.

	Researchers have explored several techniques  to overcome the challenges posed by through-the-wall (TTW) sensing \cite{17,18,ttw1,ttw2}. However, the signal variation and power attenuation caused by walls make it difficult to detect human presence accurately. Additionally, dense deployment of transceivers is required for existing TTW detection, which strikes a compelling balance between the deployment cost and detection accuracy. Our previous work \cite{ourcsichu} presented a conditional recurrent architecture-based multiple room presence detection system that uses a single access point (AP) in each room to enable multi-room presence detection. Using TTW signals can lead to severe multipaths in narrow rooms and no signal reflection in open spaces, making it crucial to extract proper characteristics when the user is located at a LoS or NLoS position for improved detection accuracy.

	To address these challenges, we propose a time-selective conditional dual-feature extraction recurrent network (\netw) for human presence detection in severe indoor environments. With the dynamic and static (\preproc) data preprocess, both dynamic and static features are extracted as the input of our network, which employs a dual gated recurrent unit (GRU) \cite{29} network for temporal, spatial, and moving feature extraction. We incorporate each GRU extractor with a convolutional neural network (CNN) \cite{30} to provide the current spatial feature, which yields complementary information as a condition for multiple room identification. A time-selection scheme is designed in our system using an attention-based method to derive the feature of variation in the room. Finally, a majority voting process is presented for enhanced detection accuracy, and a real-time experiment is implemented to evaluate our system. The main contributions of this paper are elaborated as follows.
\begin{itemize}

	\item We have proposed a CSI-based multi-room human presence detection system with a star network topology  as shown in Fig. \ref{fig:systemArchitecture} that reduces the number of APs to one per room. We also address the issues observed in our experiments, such as LoS path blocking and non-blocking, misjudgment in environments with scarce multipath, and signal attenuation in the TX room. Note that TX room is referred as the center room with transmitter as shown in Fig. \ref{fig:systemArchitecture}; whereas those rooms with receivers are named as RX rooms. Additionally, we have designed a multi-room voting process to average the probabilities of multiple transmission pairs for the TX room, which enhances the prediction performance due to signal attenuation problem.

	\item To address the problems of LoS path blocking and non-blocking as well as scarce multipath environment, we propose the \netw system which extracts significant time features for scarce multipath environments using an attention-based method. Additionally, we propose the \preproc data preprocessing method, which works in conjunction with the dual network of the \netw system to distinguish between LoS blocking path and non-blocking path by extracting dynamic and static features separately. Furthermore, we design a specific loss function with CNN in the condition part to generate the current feature of human presence.

	\item We have conducted a series of experiments to observe the characteristics of CSI from three different aspects. The first aspect is that the CSI amplitude shape is sensitive to human motion in the presence of LoS path, and vice versa. The second aspect is that the CSI amplitude variance in environments with either rich or scarce multipath can lead to misjudgement in empty rooms. The third aspect is that the distinctive CSI amplitude patterns of human presence reveal at the rooms with either TX or RX. Finally, we evaluate the effectiveness of proposed TCD-FERN in multi-room presence detection, which outperforms the other existing algorithms in open literature.

\end{itemize}

The remainder of this paper is organized as follows. In Section \ref{CHP_PRE}, we present the system architecture and preliminary of CSI. Next, the proposed system model \netw is illustrated in Section \ref{CHP_PSM}. In Section \ref{CHP_ME}, experimental setting and CSI observation are elaborated, whereas performance evaluation and comparison of proposed TCD-FERN scheme are discussed in Section \ref{CHP_ER}. Finally, the conclusion is drawn in Section \ref{CHP_CON}.

\section{System Architecture} \label{CHP_PRE}

\subsection{System Architecture}
\label{SYS_COMP}

\begin{figure}
\centering
\includegraphics[width=1\linewidth]{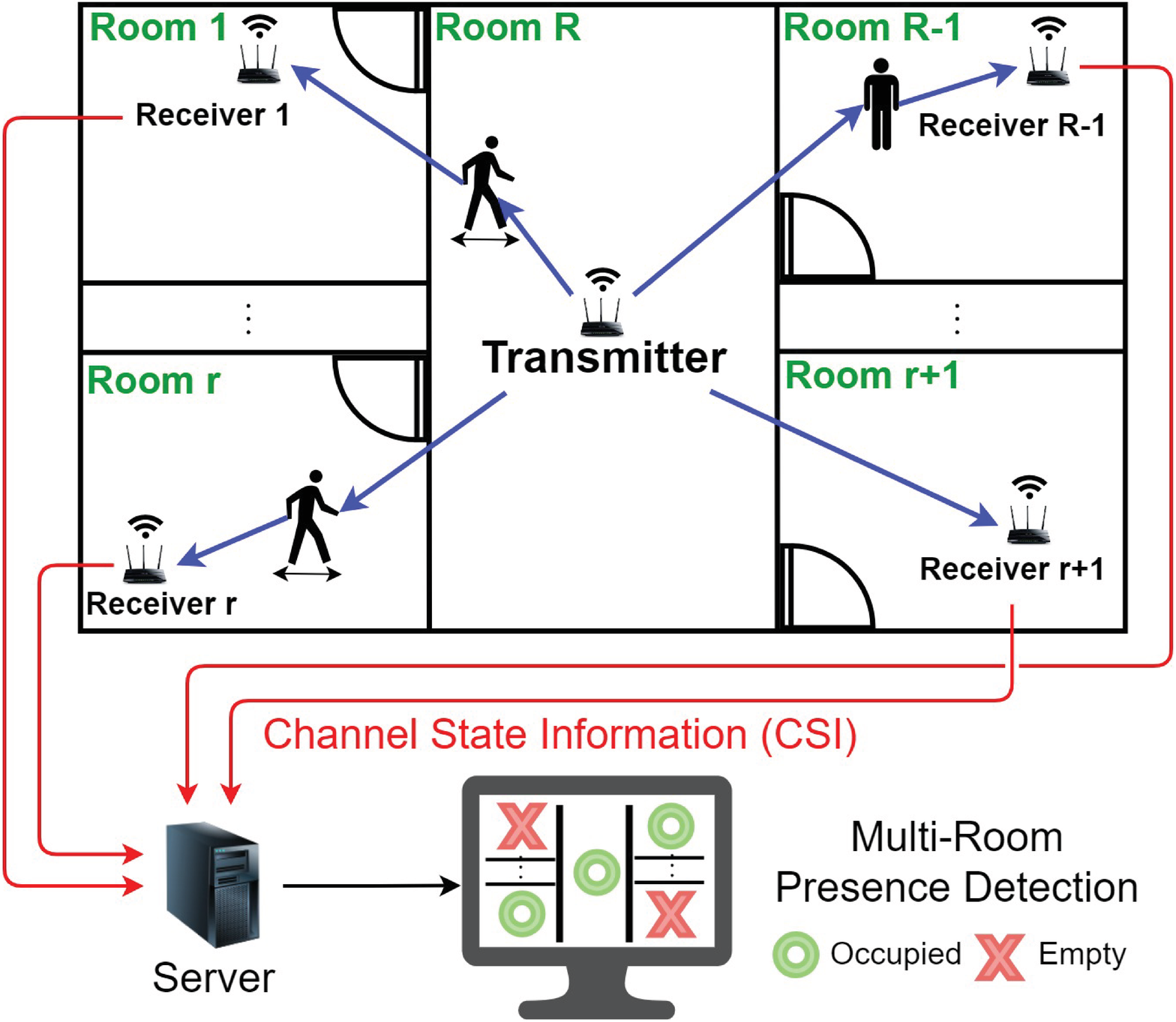}
\caption{System architecture of human presence detection in a multi-room scenario. The edge server will collect measured CSI data for presence detection.}
\label{fig:systemArchitecture}
\end{figure}

Our system is designed to detect the presence of humans in multiple rooms without the need for user-equipped devices but only commercial off-the-shelf WiFi APs. As depicted in Fig. \ref{fig:systemArchitecture}, our system architecture consists of multiple indoor rooms, containing a single transmitter and several receivers. The number of receivers deployed depends on the number of rooms being considered. For instance, if there are $R$ rooms, one transmitter is deployed in the center room, whereas there are $R-1$ receivers in the surrounding rooms. The transmitter continuously sends out RF signals to the receivers, whilst receivers transmit the collected CSI of the particular receiver channel to the server. Our algorithm is then applied to determine the presence of humans in each room.

Fig. \ref{fig:systemArchitecture} also depicts that two adjacent rooms can be measured by a single transmission pair consisting of a transmitter and a receiver. Note that such deployment is considered more applicable for realistic implementation compared to the adoption of a pair of APs in each room in most existing works. We define a transmission pair indexed by $p \in [1, P]$, where $P$ is the total number of transmission pairs. Each transmission pair analyzes individual pair of rooms, and the channel information can be sorted into four categories based on the presence of humans in each room:
\begin{itemize}
	\item \textbf{Case 1}: both rooms are empty.
	\item \textbf{Case 2}: human presence only in TX room.
	\item \textbf{Case 3}: human presence only in RX room.
	\item \textbf{Case 4}: human presence in both TX and RX rooms.
\end{itemize} 
The existence of humans in the two rooms can be determined by analyzing these cases.


\subsection{Channel State Information} 
\label{CSI}
The use of multiple input multiple output (MIMO) systems in digital wireless communication has become widespread \cite{27}. A MIMO system comprises multiple transmitting and receiving antennas, which provide abundant CSI and increase data throughput by enhancing channel diversity. In the case of WiFi routers, MIMO is combined with orthogonal frequency division multiplexing (OFDM) \cite{26}. OFDM divides the channel into orthogonal frequency bands, which do not interfere with each other, thereby enhancing system bandwidth efficiency. Additionally, it can resolve frequency-selective fading in multipath propagation environments. In typical indoor environments, WiFi signals travel from the transmitter through various paths to the receiver, with obstacles such as walls and furniture causing reflection, refraction, scattering, and diffraction. These CSI signals describe the channel properties, and each subcarrier is expressed as a complex number that contains amplitude attenuation and phase shift. Therefore, the received signal of the channel can be expressed as
\begin{align}
{y}^p_{t,q,k} = h^p_{t,q,k} x^p_{t,q,k} +{z} ^p_{t,q,k},
\end{align}
where $x^p_{t,q,k}$ and ${y}^p_{t,q,k}$ are the transmitted and received signals of the $q$-th subcarrier and $k$-th antenna pair of the $p$-th transmission pair at time $t$. $h^p_{t,q,k}$ is the channel response, and ${z} ^p_{t,q,k}$ is the additive white Gaussian noise (AWGN). The estimated CSI of the $q$-th subcarrier and $k$-th antenna pair of the $p$-th transmission pair at time $t$ is then given by
\begin{align}
\hat{h}^p_{t,q,k} = \frac{{y}^p_{t,q,k}}{x^p_{t,q,k}}.
\end{align}
Hence, the CSI matrix at the receiver of the $p$-th transmission pair at time $t$ is defined as:
\begin{align}
\boldsymbol{\hat{H}}^p_t = \begin{bmatrix} \hat{h}^p_{t,1,1}&\dots&\hat{h}^p_{t,1,k}&\dots&\hat{h}^p_{t,1,K}
\\
\hat{h}^p_{t,2,1}&\dots&\hat{h}^p_{t,2,k}&\dots&\hat{h}^p_{t,2,K}
\\
\vdots&\ddots&\vdots&\ddots&\vdots
\\
\hat{h}^p_{t,Q,1}&\dots&\hat{h}^p_{t,Q,k}&\dots&\hat{h}^p_{t,Q,K} \end{bmatrix},
\end{align}
where $Q$ and $K$ are the total number of subcarriers and antenna pairs in each transmission pair, respectively. Each element of the CSI matrix can be represented as
\begin{align}
\hat{h}^p_{t,q,k} = & \left| \hat{h}^p_{t,q,k} \right| e^{j\sin \left( \angle \hat{h}^p_{t,q,k} \right)},
\end{align}
where $\left| \hat{h}^p_{t,q,k} \right|$ is the amplitude response, whilst $\angle \hat{h}^p_{t,q,k}$ is the phase response of the $q$-th subcarrier and $k$-th antenna pair of the $p$-th transmission pair at time $t$. In our system, we focus on the amplitude response of CSI and discard the phase of $\hat{h}^p_{t,q,k}$. The normalized element of the CSI matrix at time $t$ can be represented as
\begin{align}
\tilde{h}^p_{t,q,k} = \frac{\left| \hat{h}^p_{t,q,k} \right| - \min\limits_{l} \left| \hat{h}^p_{t,l,k}\right| }{\max\limits_{l}\left| \hat{h}^p_{t,l,k}\right| - \min\limits_{l}\left| \hat{h}^p_{t,l,k}\right|}, \forall l = 1,\dots,Q .
\label{eq:norm_ele}
\end{align}
The normalized CSI matrix at time $t$ is
\begin{align}
\boldsymbol{\tilde{H}}^p_t = \Big\lbrace\tilde{h}^p_{t,q,k}{\Big|} \forall q=1, \dots, Q \text{ and } \forall k=1, \dots, K\Big\rbrace.
\label{eq:norm_mat}
\end{align}
We conduct normalization to eliminate power variations caused by the distinct power generated by the OFDM channel of APs at different time instants, which causes non-uniformity of CSI data. Note that this variation is not generated by the environmental changes. Additionally, we aim to exploit the amplitude shape because our observations indicate that the shape patterns of CSI are specific for different presence or environment changes. Furthermore, OFDM multipath provides diversity, so normalization is executed on each antenna pair independently.

\section{Proposed WiFi-CSI Presence Detection System} \label{CHP_PSM}

\begin{figure}
\centering
\includegraphics[width=1\linewidth]{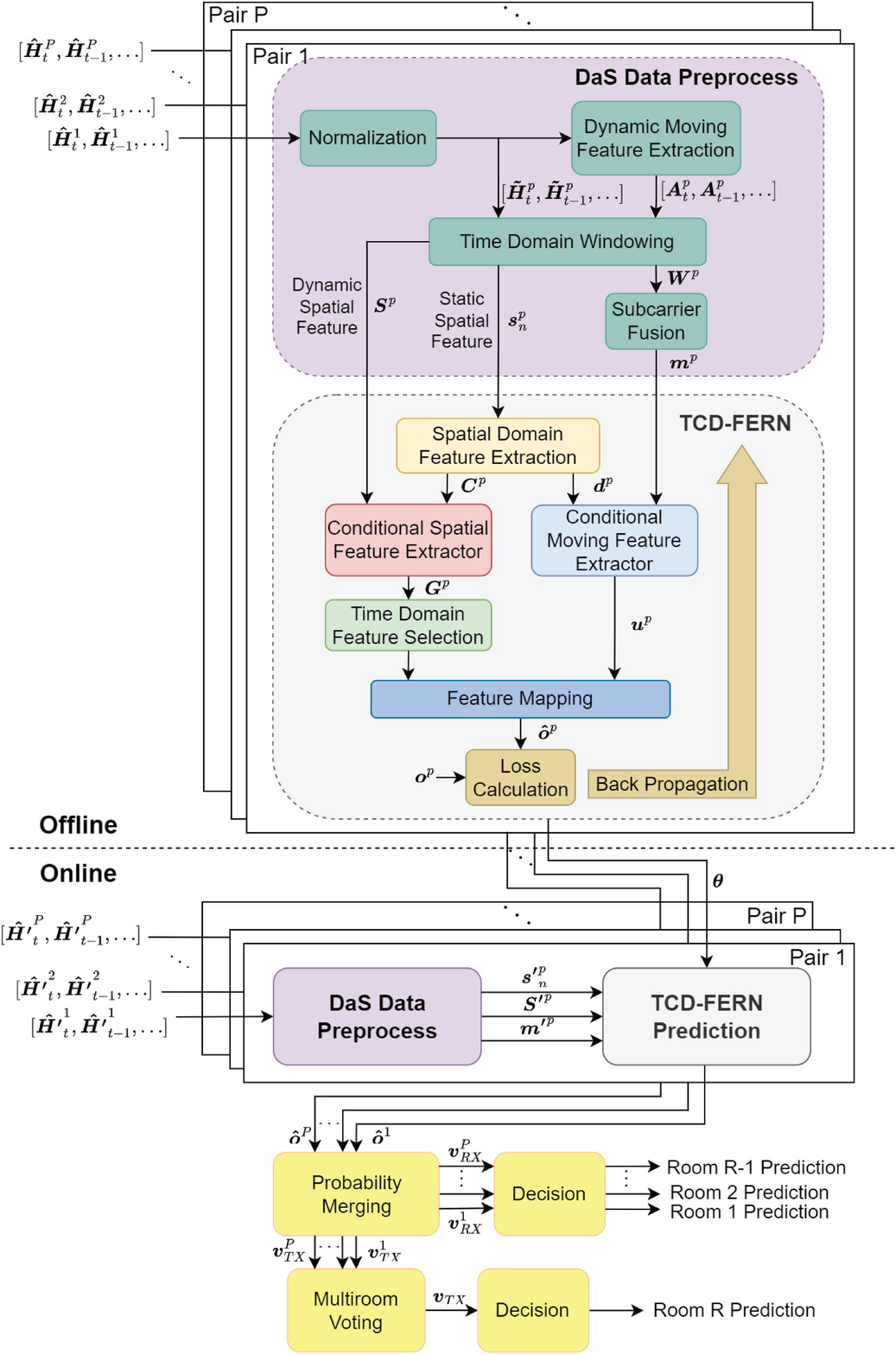}
\caption{Schematic diagram of the proposed WiFi-based human presence detection system.}
\label{fig:blockDiagram}
\end{figure}

Our proposed system, as shown in \fig \ref{fig:blockDiagram}, can be divided into two parts: the offline part and online part. In the offline stage, we preprocess the received data for each RX by conducting normalization on the input data in \preproc data preprocess. We then extract dynamic moving features from the normalized data and apply windowing to both the normalized data and dynamic moving features. The windowed normalized CSI is a dynamic spatial feature, while the windowed dynamic moving feature is used for subcarrier fusion. Additionally, we utilize the static spatial feature, which is the current CSI amplitude, as a conditional input. In \netw, the dynamic spatial feature and the dynamic moving feature after subcarrier fusion are used as input for the conditional spatial feature extractor and conditional moving feature extractor, respectively. The spatial domain feature extracted from the static spatial feature is fed as a condition to both extractors. Next, the conditional spatial feature is exploited with time domain feature selection, and the output is combined with the conditional moving feature for feature mapping. Finally, we obtain the parameters of our model by calculating the constant loss.

In the online phase, we conduct \preproc data preprocess again for the input CSI data, and acquire the static spatial feature, dynamic spatial feature, and dynamic moving feature. Then, we input these features into the \netw prediction model trained in the offline phase. Ultimately, we execute a voting scheme with the output of the \netw model to predict the human presence for each room. In the following subsections, we discuss each functional block in detail and explain their significance in accomplishing multi-room human presence detection.

\subsection{Data Preprocess}
In this subsection, we elaborate each functional block of \preproc data preprocess, which includes normalization, dynamic moving feature extraction, windowing, and subcarrier fusion. The purpose of \preproc data preprocess is to extract both dynamic and static features of human presence while reducing the dimension of the CSI data.

\subsubsection{Normalization}
As shown in \fig \ref{fig:blockDiagram}, the first step after receiving the CSI data $\boldsymbol{\hat{H}}^p_t$ of $p$-th transmission pair at time $t$ is normalization. The normalization procedure is explained in subsection \ref{CSI}, and we obtain the normalized CSI of each element by calculating (\ref{eq:norm_ele}). We then use this normalized CSI matrix denoted as $\boldsymbol{\tilde{H}}^p_t$ in (\ref{eq:norm_mat}) for all subcarriers and antenna pairs at time $t$ of $p$-th transmission pair. All subsequent processes in our system use this normalized CSI matrix $\boldsymbol{\tilde{H}}^p_t$ to extract more precise CSI features. We also consider the normalized CSI matrix $\boldsymbol{\tilde{H}}^p_t$ as a spatial feature in our system, as the CSI amplitude shape after normalization is significant and contains features of both the environment and human presence.

\subsubsection{Dynamic Moving Feature Extraction}
In addition to spatial features, human motion in a room can also cause changes in the CSI amplitude pattern. Therefore, we extract moving features in this block using the amplitude shape trend (AST) approach, which can be represented as
\begin{align}
a^p_{t,q,k} = \tilde{h}^p_{t,q,k} - \tilde{h}^p_{t-1,q,k} ,
\end{align}
where $a^p_{t,q,k}$ is the element of the dynamic moving feature of the $q$-th subcarrier and $k$-th antenna pair of the $p$-th transmission pair at time $t$. The feature at time $t$ is obtained by subtracting the previous normalized CSI amplitude at time $t-1$ from the present normalized CSI amplitude of time $t$. The outcome of the AST approach displays the changes in the CSI amplitude shape of consecutive CSI amplitude data. If the difference in the two consecutive CSI amplitudes is large, resulting in a larger $a^p_{t,q,k}$, it indicates that there can exist people in the detected room. Conversely, if the difference is small, resulting in a small $a^p_{t,q,k}$, it can be an empty room without any people. The AST approach output for all $Q$ subcarriers and $K$ antenna pairs is denoted as
\begin{align}
	\boldsymbol{A}^p_t =\boldsymbol{\tilde{H}}^p_t  - \boldsymbol{\tilde{H}}^p_{t-1} = \begin{bmatrix} a^p_{t,1,1}&a^p_{t,1,2}&\dots&a^p_{t,1,K}\\a^p_{t,2,1}&a^p_{t,2,2}&\dots&a^p_{t,2,K}\\\vdots&\vdots&\ddots&\vdots\\a^p_{t,Q,1}&a^p_{t,Q,2}&\dots&a^p_{t,Q,K} \end{bmatrix} .
\end{align}
Therefore, we use this CSI amplitude shape change of successive times, which is $\boldsymbol{A}^p_t$ of the $p$-th transmission pair at time $t$, as our dynamic moving feature.

\subsubsection{Time Domain Windowing}
After collecting the spatial feature $[ \boldsymbol{\tilde{H}}^p_t,\boldsymbol{\tilde{H}}^p_{t-1},\dots]$ and moving feature $[ \boldsymbol{A}^p_t,\boldsymbol{A}^p_{t-1},\dots ]$, we divide them into several windows with window size $\tau$. This process of windowing is important because it provides the variation of each subcarrier in a period of time, which is a key characteristic of time-dependent data. We conduct windowing by dividing the received feature vectors with a sliding window, and it is performed on both spatial and moving features separately. In the \textbf{spatial} part, we perform windowing on the normalized CSI amplitude. The vector of spatial samples for the $p$-th transmission pair at time $t$ can be represented as
\begin{align} \label{stas}
	 \boldsymbol{s}^p_{t} = \left[ \tilde{h}^p_{t,1,1},\tilde{h}^p_{t,2,1},\dots,\tilde{h}^p_{t,q,k},\dots,\tilde{h}^p_{t,Q,K} \right],
\end{align}
where the elements of this vector are the normalized CSI amplitudes of different combinations of subcarriers and antenna pairs. Comparing the elements of vector $\boldsymbol{s}^p_{t}$ with the elements of matrix $\boldsymbol{\tilde{H}}^p_t$ in (\ref{eq:norm_mat}), it can be easily seen that $\boldsymbol{s}^p_{t}$ is the transformation of $\boldsymbol{\tilde{H}}^p_t$ from a matrix into a vector. Then, the matrix of samples with a window size of $\tau$ can be expressed as
\begin{align}
	\boldsymbol{S}^p = \left[ \boldsymbol{s}^p_{n-\tau},\boldsymbol{s}^p_{n-\tau+1},\dots,\boldsymbol{s}^p_{n} \right]^{T} , \quad \forall n=1,\dots,N ,
\end{align}
where $N$ is the number of samples collected. The windowed spatial feature $\boldsymbol{S}^p$ is utilized as the \textit{dynamic spatial feature} for our following \netw, since the amplitude shape variation in a period provides spatial information. In addition, we use the last vector of the matrix $\boldsymbol{S}^p$, which is $\boldsymbol{s}^p_{n}$, as the \textit{static spatial feature}. Since $\boldsymbol{s}^p_{n}$ comprises the CSI amplitude waveform of sample $n$, it contains the information of the current time. For the \textbf{moving} part, we apply windowing to the \textit{dynamic moving feature} $\boldsymbol{A}^p_t$ extracted in the previous block. The vector of moving samples for the $p$-th transmission pair at time $t$ can be expressed as
\begin{align}
	 \boldsymbol{w}^p_{t} = \left[ a^p_{t,1,1},a^p_{t,2,1},\dots,a^p_{t,q,k},\dots,a^p_{t,Q,K} \right] ,
\end{align}
where the elements of this vector consist of the elements in $\boldsymbol{A}^p_t$. Similar to the spatial part, $\boldsymbol{w}^p_{t}$ is the transformation of $\boldsymbol{A}^p_t$ from a matrix into a vector. We then divide the dynamic moving feature into a matrix of samples with a window size of $\tau$, which can be expressed as
\begin{align}
	\boldsymbol{W}^p = \left[ \boldsymbol{w}^p_{n-\tau},\boldsymbol{w}^p_{n-\tau+1},\dots,\boldsymbol{w}^p_{n} \right]^{T} , \;\;\;\forall n=1,\dots,N.
\end{align}
Note that the windowed dynamic moving feature, represented by the matrix $\boldsymbol{W}^p$, provides the variance of waveform over time, which reflects the magnitude of amplitude change.

\subsubsection{Subcarrier Fusion}
\label{SF}
After windowing the dynamic moving feature $\boldsymbol{W}^p$, the next step is to perform subcarrier fusion on $\boldsymbol{W}^p$. This is accomplished by taking the mean of all subcarriers in a sample:
\begin{align}
m^p_{n} = \frac{1}{Q \cdot K} \sum^{Q}_{i=1} \sum^{K}_{l=1} {a^p_{t,i,l}} ,
\end{align}
where $m^p_{n}$ is the $n$-th element of the dynamic moving feature after subcarrier fusion for $p$-th transmission pair, $Q$ is the number of subcarriers, and $K$ is the number of antenna pairs. The windowed dynamic moving feature $\boldsymbol{W}^p$ represents the variation of successive data, which is generated by the transition of the environment, resulting in different magnitudes or shapes. Since each subcarrier can be considered as an independent channel, when occupancy occurs, some channels are affected by the human while others are not. For instance, when a person is walking in a NLoS area, some multipath signals will not be reflected by the person, but rather by the wall on the opposite side of the room. To provide mutative information, we extract dynamic moving features. However, incorporating data from all subcarriers causes complicated predictions due to the overabundance of information and increases the risk of overfitting. Hence, we take the average of these variables, which provides a more objective variation of each sample and reduces the computational complexity. After performing subcarrier fusion on each sample, the vector of fused dynamic moving feature of $p$-th transmission pair is represented as:
\begin{align}
\boldsymbol{m}^p = \left[ {m}^p_{n-\tau},{m}^p_{n-\tau+1},\dots,{m}^p_{n} \right], \quad \forall n=1,\dots,N.
\end{align}
The vector $\boldsymbol{m}^p$ serves as the representation of the moving feature for the following \netw algorithm.

\subsection{\netw}

\begin{figure}
\centering
\includegraphics[width=1\linewidth]{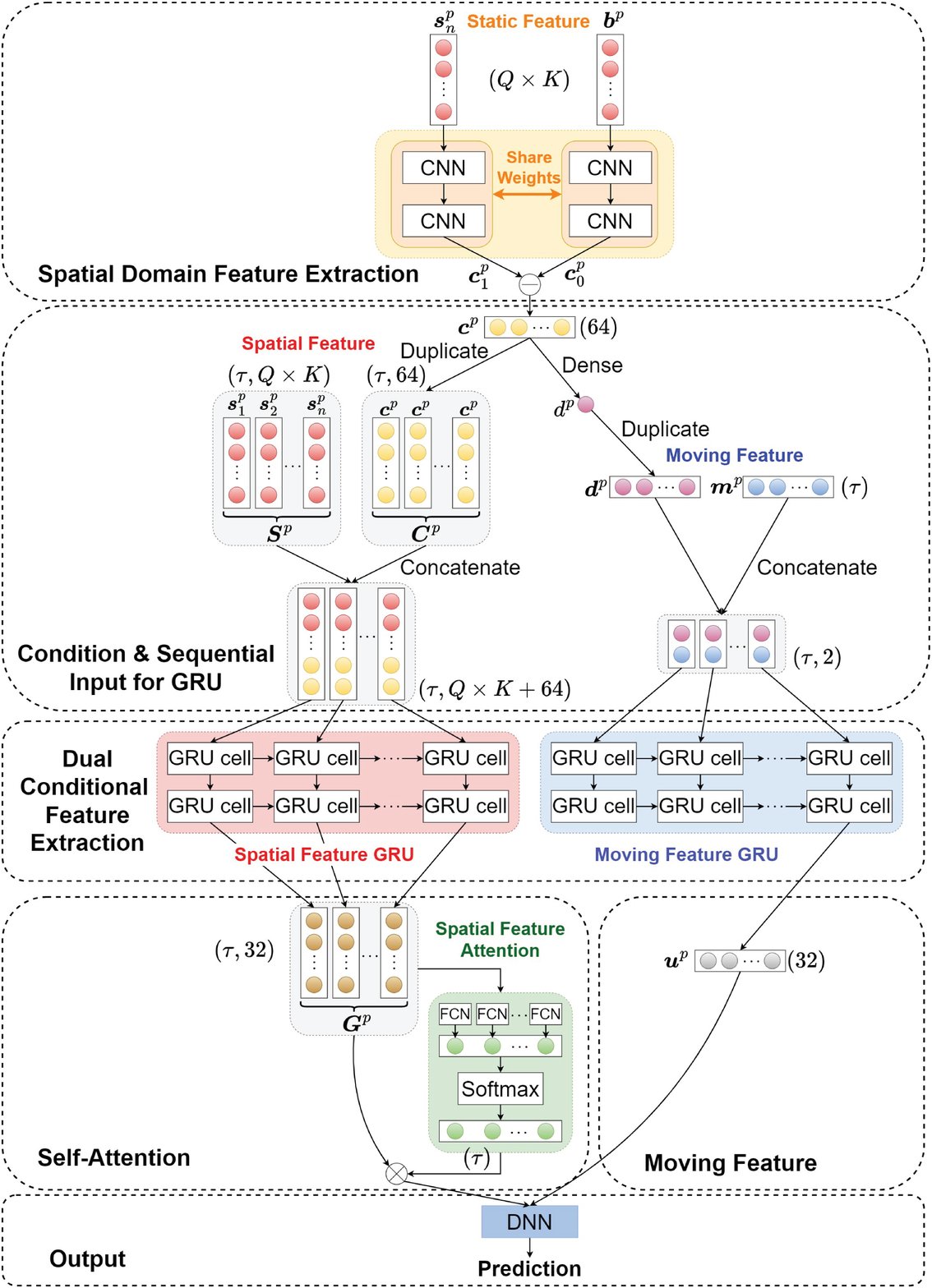}
\caption{Proposed architecture of time-selective conditional dual recurrent network.}
\label{fig:Network}
\end{figure}

After obtaining $\boldsymbol{S}^p$, $\boldsymbol{m}^p$, and $\boldsymbol{s}^p_{n}$ from the \preproc data preprocess, we introduce our \netw algorithm as shown in \fig \ref{fig:blockDiagram}. We explain each block in this subsection, and the corresponding flow diagram of the proposed network is illustrated in \fig \ref{fig:Network}. We start by concatenating the static spatial feature $\boldsymbol{S}^p$ and dynamic moving one $\boldsymbol{m}^p$. Then, we perform dual conditional feature extraction. For the spatial part, we apply self-attention with time selection. The output, along with the moving feature, is input to a deep neural network for feature mapping to provide the final prediction. In the following subsections, we first introduce the main dual network, followed by the spatial domain feature extraction. Finally, we present time selection and feature mapping.

\subsubsection{Dual Recurrent Neural Network}
\label{dualGRU}
The dual conditional feature extraction block, as depicted in \fig \ref{fig:Network}, comprises two networks each with two layers of GRU cells, forming a dual network. Although GRU and long-short-term memory (LSTM) \cite{25} units are similar in that they both extract characteristics of time with sequential data and resolve the gradient vanishing problem, GRU does not have separate memory cells. With fewer gates than LSTM, GRU has fewer parameters and reduces both the complexity and training time. To elaborate a little further, LSTM can tackle large sequence of data; however, human presence detection usually requires a short time-window size to prevent compellingly high latency of providing the final results. In our system, we design a dual recurrent neural network instead of a common single network because the two inputs have different features to extract, as mentioned above. If we consider both $\boldsymbol{S}^p$ and $\boldsymbol{m}^p$ in a single GRU network, the extraction from the neural network may be dominated by dynamic moving features, since little signal changes for the static cases may be indistinguishable from each other \cite{cronos}. Therefore, we design a dual GRU network to highlight the strengths of respective static and dynamic features for classification.

In our dual network architecture, one GRU is responsible for conditional spatial feature extraction, while the other focuses on conditional moving feature extraction. The conditional spatial feature extractor takes the spatial feature $\boldsymbol{S}^p$ as its main input, which represents the amplitude shape of the CSI and is used to differentiate between the presence of a static human and an empty room. To detect the motion of humans, we use the dynamic moving feature $\boldsymbol{m}^p$, which reflects the relationship between two consecutive CSI amplitudes. When people stay at LoS, the dynamic moving feature captures the difference between two successive CSI amplitudes, which is not affected by the shape of CSI amplitude especially the shape of the LoS path signal. On the other hand, when people are in the NLoS area, the spatial feature can effectively capture the fluctuation caused by relatively severe multipath in narrow rooms. Nevertheless, in open spaces with scarce signal reflections, observing the variation between CSI amplitudes will be more effective. Therefore, our system employs both dynamic spatial feature and dynamic moving feature to achieve better performance.

\subsubsection{Spatial Domain Feature Extraction}

The input for the dual conditional network not only comprises of $\boldsymbol{S}^p$ and $\boldsymbol{m}^p$ discussed in the previous subsection, but also includes features extracted from the spatial domain. The spatial domain features consist of the current CSI amplitude features of each input window, $\boldsymbol{S}^p$ and $\boldsymbol{m}^p$, which are extracted using two layers of CNN. The CNN applies convolution on input graphs with filters and outputs feature maps. Additionally, the input includes the static spatial feature, $\boldsymbol{s}^p_{n}$, which is extracted from preprocessed data. The output of each CNN layer can be expressed as
\begin{align} \label{c1}
	\boldsymbol{c}^p_{1,\gamma} \left[ \epsilon \right] = \left( \boldsymbol{s}^p_{n}*\Phi^p_{1,\gamma} \right) \left[ \epsilon \right] , 
\end{align}
where $\gamma$ represents the index number of the CNN layer, $\Phi^p_{1,\gamma}$ denotes the kernel of the $\gamma$-th layer of the $p$-th transmission pair, $*$ represents the convolution operation, and $\epsilon$ is the index of a data sample. The input for the first layer is $\boldsymbol{s}^p_{n}$, while the subsequent layers use the output from the previous layer, denoted as $\boldsymbol{c}^p_{1,\gamma-1}$, as input. Finally, in the last layer, we obtain the spatial domain feature $\boldsymbol{c}^p_1$, which contains the features from the CSI shapes of the current data and can be regarded as a multipath feature.

\begin{figure}
\centering
\subfigure[]{\includegraphics[width=1\linewidth]{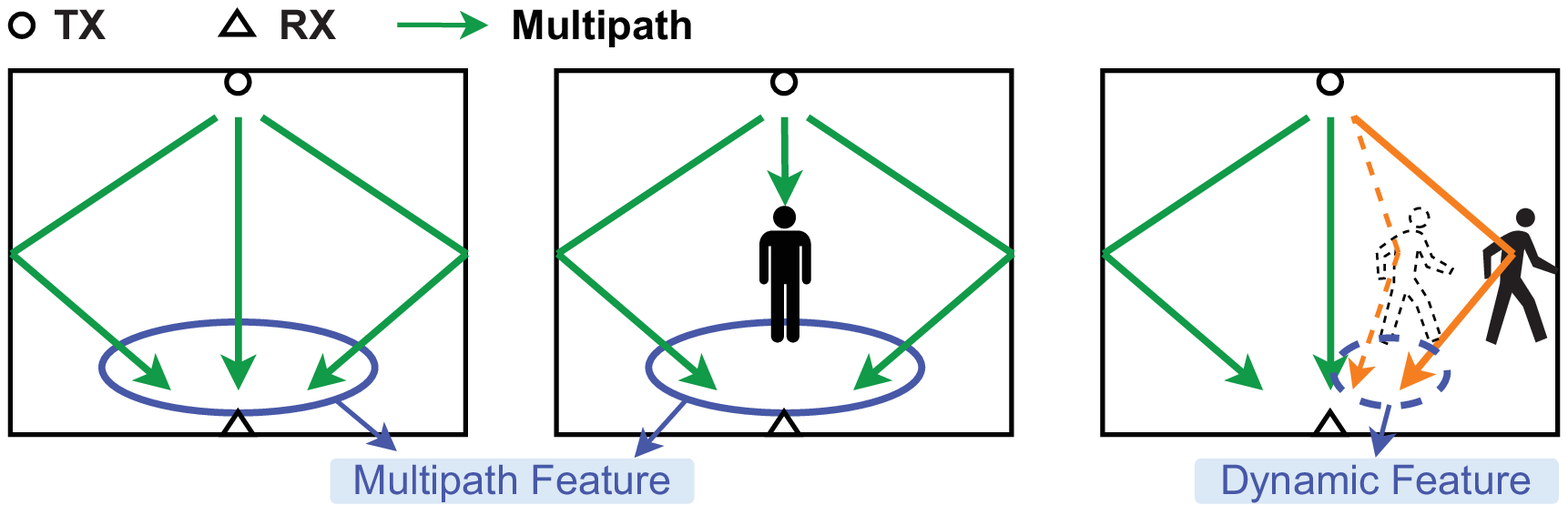} \label{fig:cond1}	}
\quad
\subfigure[]{\includegraphics[width=1\linewidth]{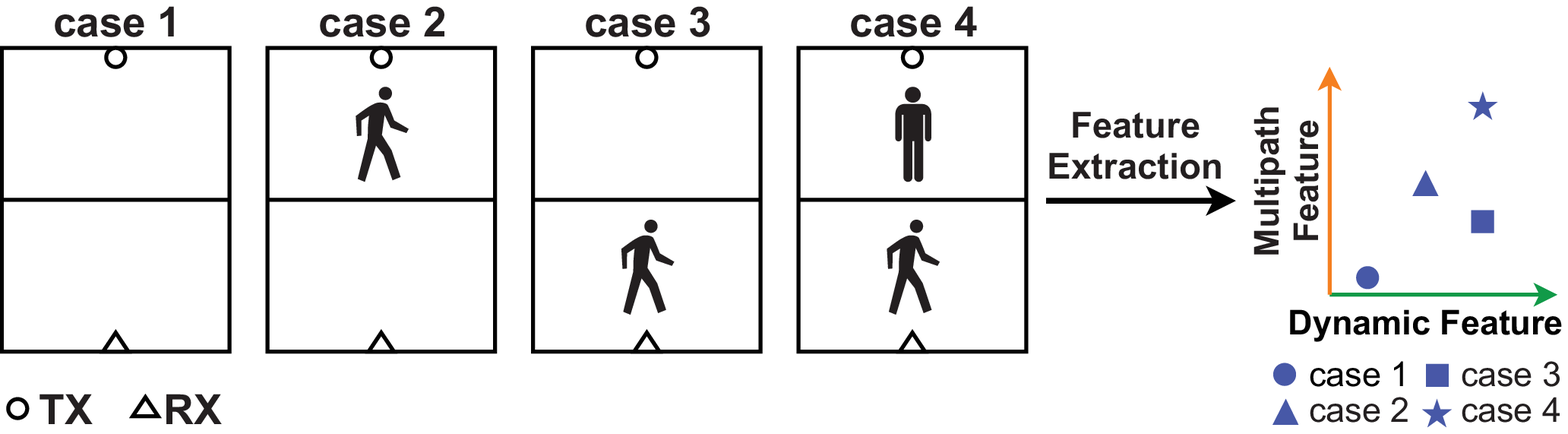} \label{fig:cond2}}
\caption{Visual explanation of the spatial domain feature. (a) Illustration of multipath feature and dynamic feature. (b) Two dimensional feature for four cases of human presence in two-room scenarios.}
\label{ConditionExplain}
\end{figure}

As shown in Fig. \ref{fig:cond1}, unlike the dynamic feature generated by human motion, the multipath feature contains mainly the environmental information and variations or the environment affected by humans. Since this is additional information for the GRU networks, it can be considered as a conditional input. Using $\boldsymbol{c}^p_1$ as the condition and inputting it to the dual recurrent neural network along with $\boldsymbol{S}^p$ and $\boldsymbol{m}^p$, the network is forced to consider the extra relationship between the current data and the $\tau$ samples in the windows of $\boldsymbol{S}^p$ and $\boldsymbol{m}^p$, in addition to abstracting the relationship between data in a window. Therefore, the output of the dual recurrent neural network grouped into one cluster may be divided into several subclusters after the complement of the condition. With the increase in the number of features, this novel design can appropriately distinguish between different rooms. 

For instance, as shown in Fig. \ref{fig:cond2}, in cases 2 and 3, there is only one person walking in one of the rooms. In the original model without the conditional input, GRU extracts the feature of successive time duration by adjusting weights. However, the limitation of the block length, which is the given window size $\tau$, results in the neglect of some fine features after training in hidden layers with massive neurons. Since the people in cases 2 and 3 have the same dynamic feature, the multipath feature implicitly containing the human position information helps in better differentiating between different rooms. The difference from case 4 to case 3 in Fig. \ref{fig:cond2} is that there is an additional person standing still in the TX room while another person walks in the same pattern in the RX room. In these two cases, it is evident that the dynamic feature generated by the person walking in the RX room will provide more significant feature information compared to static spatial feature. After the dual recurrent neural network feature extraction, the dominating dynamic feature, which is the feature of time, makes it difficult to identify the difference between cases 3 and 4. With our novel design to include the current CSI amplitude shape from multipath effect, we can provide spatial information that is affected by the humans presence in the environment. Considering this condition can further ensure that the previous signals refer to the current state. Therefore, when the dynamic feature is identical, the multipath feature that captures environmental changes can further be served as a conditional input to precisely classify different cases of human presence in multi-room scenarios. As shown in the right of Fig. \ref{fig:cond2}, the two dimensional plot with both the dynamic and multipath features can effectively enhance the prediction performance by distinguishing subtle feature difference.

Additionally, we aim to impose a condition based on the feature of current spatial fluctuation, i.e., to signify the shape fluctuation of the room when people are present. Therefore, as shown in the block of Spatial Domain Feature Extraction of Fig. \ref{fig:Network}, we employ spatial sample of case 1 $\boldsymbol{b}^p$ as a reference, which represents the environmental shape of an empty room. Note that $\boldsymbol{b}^p$ is acquired identically to $\eqref{stas}$ with the same dimension. Similarly, we input $\boldsymbol{b}^p$ into a CNN network that shares the same weights as the CNN network extracting features of $\boldsymbol{s}^p_{n}$, where each layer of CNN can be expressed as
\begin{align} \label{c0}
	\boldsymbol{c}^p_{0,\gamma}\left[ \epsilon \right] = \left( \boldsymbol{b}^p*\Phi^p_{0,\gamma} \right) \left[ \epsilon \right] , 
\end{align}
where $\boldsymbol{b}^p$ is the input for the first layer, whereas the rest of CNN layers has the previous output $\boldsymbol{c}^p_{0,\gamma-1}$ as input. As a result, we obtain the output environment feature, denoted as $\boldsymbol{c}^p_0$, in the final layer. Note that two original signals of $\boldsymbol{s}^p_n$ (static spatial features) and $\boldsymbol{b}^p$ (both rooms empty) are loosely coupled, since static cases consist of both rooms empty and human presence at LoS as well as at NLoS. We intend to extract the case of either one of two rooms with human presence from the case of both empty rooms.

Next, to obtain the current spatial domain variance caused by humans, we subtract the environment feature from the present spatial domain feature as
\begin{align} \label{condii}
	\boldsymbol{c}^p = \boldsymbol{c}^p_1-\boldsymbol{c}^p_0 ,
\end{align}
where $\boldsymbol{c}^p_1$ and $\boldsymbol{c}^p_0$ are the features obtained after spatial domain feature extraction in $\eqref{c1}$ and $\eqref{c0}$, respectively. To input this spatial domain feature $\boldsymbol{c}^p$ as a condition with $\boldsymbol{S}^p$ in the spatial part, we replicate it $\tau$ times, represented as $\boldsymbol{C}^p$, where $\boldsymbol{C}^p = [\boldsymbol{c}^p,\boldsymbol{c}^p ,\dots,\boldsymbol{c}^p]$. For the moving part, the spatial domain feature is $\boldsymbol{d}^p = [{d}^p,{d}^p,\dots,{d}^p]$, where ${d}^p$ is the feature obtained after applying the dense layer to $\boldsymbol{c}^p$, and $\boldsymbol{d}^p$ is the vector obtained after duplicating ${d}^p$ by $\tau$ times.

Furthermore, considering the four cases that we are trying to classify, we aim to provide variant $\boldsymbol{c}^p$ as a condition to improve the performance. Therefore, we use a conditional loss as
\begin{align}
	l^p_{cond} &= \frac{1}{N} \sum^{N}_{n=1} \left[ \mathbbm{1}(y=0) \boldsymbol{\nu}^2_n \right. \notag\\
	& \left. + \mathbbm{1}(y\neq 0)\max \left( y \times margin-\boldsymbol{\nu}_n, 0 \right)^2 \right] ,
\label{eq:con_loss}
\end{align}
where $\boldsymbol{\nu}_n = \left| \boldsymbol{c}^p_1-\boldsymbol{c}^p_0 \right|$ represents the distance between the two features. $N$ is the total number of considered samples, $margin$ is a hyperparameter, and the indicator functions $\mathbbm{1}\left(y=0\right)$ and $\mathbbm{1}\left(y\neq 0\right)$ denote the circumstances when the label represents an empty room or not. Thus, when the inputs $\boldsymbol{s}^p_{n}$ and $\boldsymbol{b}^p$ have a large difference, the dissimilarity of the feature after spatial domain feature extraction would be larger. The intuition behind this loss function is that the distance $\boldsymbol{\nu}_n$ should be smaller in case 1 since they both represent an empty room with noises as the main sources of difference. For the other cases, we wish the dissemblance of the features, compared to the environment spatial feature, to vary with the degree of the CSI amplitude variance in each case.

\begin{figure}
\centering
\includegraphics[width=.9\linewidth]{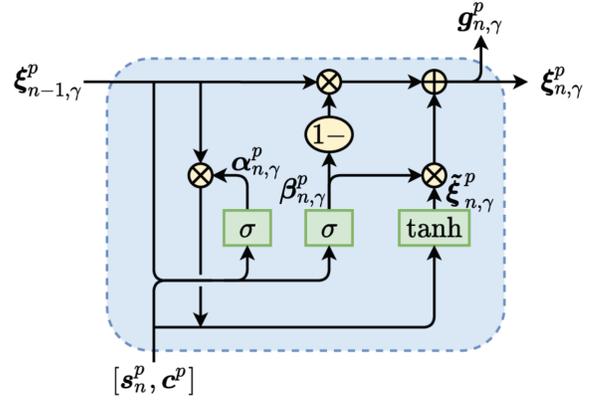}
\caption{The architecture of a single GRU cell.}
\label{fig:gru}
\end{figure}

After inputting $\boldsymbol{S}^p$ with the condition $\boldsymbol{C}^p$ in the spatial part into the conditional spatial feature extractor, the dual recurrent neural network extracts the conditional spatial feature for each sample, which is combined with the structure of a GRU unit shown in Fig. \ref{fig:gru} as
\begin{align}
	 \boldsymbol{\alpha}^p_{n,\gamma} &= \sigma \left( \Omega_{\alpha,\gamma} \cdot \left[ \boldsymbol{\xi}^p_{n-1,\gamma}, \boldsymbol{s}^p_{n}, \boldsymbol{c}^p \right] \right), \\  
	 \boldsymbol{\beta}^p_{n,\gamma} &= \sigma \left( \Omega_{\beta,\gamma} \cdot \left[ \boldsymbol{\xi}^p_{n-1,\gamma}, \boldsymbol{s}^p_{n}, \boldsymbol{c}^p \right] \right),  \\   
	 \boldsymbol{\tilde{\xi}}^p_{n,\gamma} &= \tanh \left( \Omega_{\tilde{\xi},\gamma} \cdot \left[ \boldsymbol{\alpha}^p_{n,\gamma} * \boldsymbol{\xi}^p_{n-1,\gamma}, \boldsymbol{s}^p_{n}, \boldsymbol{c}^p \right] \right), \\   
	 \boldsymbol{\xi}^p_{n,\gamma} &= \left( 1 - \boldsymbol{\beta}^p_{n,\gamma} \right) * \boldsymbol{\xi}^p_{n-1,\gamma} + \boldsymbol{\beta}^p_{n,\gamma} * \boldsymbol{\tilde{\xi}}^p_{n,\gamma} ,  \\  
	 \boldsymbol{g}^p_{n,\gamma} &= \sigma \left( \Omega_{g,\gamma} \cdot \boldsymbol{\xi}^p_{n,\gamma} \right) ,
\end{align}
where $\Omega_{\alpha,\gamma}$, $\Omega_{\beta,\gamma}$, $ \Omega_{\tilde{\xi},\gamma}$ and $ \Omega_{g,\gamma}$ are neural weights, $\boldsymbol{\alpha}^p_{n,\gamma}$ is the reset gate output, $\boldsymbol{\beta}^p_{n,\gamma}$ is the update gate output, and $\boldsymbol{\tilde{\xi}}^p_{n,\gamma}$ is the candidate hidden state. Notation $\boldsymbol{\xi}^p_{n,\gamma}$ is the hidden state of the $\gamma$-th layer, whilst $\boldsymbol{g}^p_{n,\gamma}$ is the output with the previous hidden state output $\boldsymbol{\xi}^p_{n-1,\gamma}$. The spatial feature $\boldsymbol{s}^p_{n}$ and the condition feature $\boldsymbol{c}^p$ are used as inputs for the first layer, while the subsequent layers take the previous hidden layer output $\boldsymbol{\xi}^p_{n-1,\gamma}$ as well as the output of the previous layer $\boldsymbol{g}^p_{n,\gamma-1}$ as inputs. $\sigma(\cdot)$ and $\tanh(\cdot)$ are the sigmoid and hyperbolic tangent activation function, respectively. Hence, the output of the final layer can be presented as
\begin{align}
	 \boldsymbol{G}^p  = \left[ \boldsymbol{g}^p_{n-\tau}, \boldsymbol{g}^p_{n-\tau+1}, \dots, \boldsymbol{g}^p_{n} \right]^T ,
\end{align}
which can be formulated in the conditional probability form $ \boldsymbol{G}^p = f \left( \boldsymbol{S}^p {\big|} \boldsymbol{C}^p \right)$. Similarly, the output of conditional moving feature extractor with input $\boldsymbol{m}^p$ and condition $\boldsymbol{d}^p$ can be expressed as $ \boldsymbol{u}^p = f \left( \boldsymbol{m}^p {\big|} \boldsymbol{d}^p \right)$.

\begin{table*}[!t]
\footnotesize
\centering
\caption{Parameter Setting of Neural Network in TCD-FERN}
\begin{tabular}{|llcl|}
\hline
\multicolumn{4}{|c|}{\textbf{Spatial Domain Feature Extraction}} \\ \hline
\multicolumn{1}{|c|}{Network Layer} & \multicolumn{1}{c|}{Symbol} & \multicolumn{1}{c|}{Dimension} & \multicolumn{1}{c|}{Note} \\ \hline
\multicolumn{1}{|l|}{Input} & \multicolumn{1}{l|}{($Q\times K$, 1)} & \multicolumn{1}{c|}{(224, 1)} & Data Input \\ \hline
\multicolumn{1}{|l|}{1D-CNN} & \multicolumn{1}{l|}{($Q\times K-$\#Kernel$+1$, \#1st filter)} & \multicolumn{1}{c|}{(222, 64)} & \begin{tabular}[c]{@{}l@{}}Filter size: 64, Kernel size: 3, Stride: 1, Dropout: 0.2\\ Activation funtion: tanh\end{tabular} \\ \hline
\multicolumn{1}{|l|}{1D-CNN} & \multicolumn{1}{l|}{($Q\times K-$2(\#Kernel$-1$), \#2nd filter)} & \multicolumn{1}{c|}{(220, 32)} & \begin{tabular}[c]{@{}l@{}}Filter size: 64, Kernel size: 3, Stride: 1, Dropout: 0.2\\ Activation funtion: tanh\end{tabular} \\ \hline
\multicolumn{1}{|l|}{Flatten} & \multicolumn{1}{l|}{($Q\times K-$2(\#Kernel$-1$) $\times$ \#2nd filter)} & \multicolumn{1}{c|}{(7040)} & \multicolumn{1}{c|}{} \\ \hline
\multicolumn{1}{|l|}{Dense} & \multicolumn{1}{l|}{($| \boldsymbol{c}^p |$)} & \multicolumn{1}{c|}{(64)} & \multicolumn{1}{c|}{} \\ \hline\hline
\multicolumn{4}{|c|}{\textbf{Condition and Sequential Input for GRU}} \\ \hline
\multicolumn{1}{|c|}{Network Layer} & \multicolumn{1}{c|}{Symbol} & \multicolumn{1}{c|}{Dimension} & \multicolumn{1}{c|}{Note} \\ \hline
\multicolumn{1}{|l|}{\begin{tabular}[c]{@{}l@{}}Input from \\ previous block\end{tabular}} & \multicolumn{1}{l|}{(1, $| \boldsymbol{c}^p |$)} & \multicolumn{1}{c|}{(1, 64)} & Input from Spatial Domain Feature Extraction \\ \hline
\multicolumn{1}{|l|}{Duplicate} & \multicolumn{1}{l|}{($\tau$, $| \boldsymbol{c}^p |$)} & \multicolumn{1}{c|}{(50, 64)} & Static feature \\ \hline
\multicolumn{1}{|l|}{\begin{tabular}[c]{@{}l@{}}Concatenate \\ (Static output)\end{tabular}} & \multicolumn{1}{l|}{($\tau$, $K\times Q + |\boldsymbol{c}^p|$)} & \multicolumn{1}{c|}{(50, 288)} & \begin{tabular}[c]{@{}l@{}}Static features from number of subcarriers (50, 224) \\ and length of latent (50, 64)\end{tabular} \\ \hline
\multicolumn{1}{|l|}{Dense} & \multicolumn{1}{l|}{(1)} & \multicolumn{1}{c|}{(1)} & Moving feature \\ \hline
\multicolumn{1}{|l|}{Duplicate} & \multicolumn{1}{l|}{($\tau$, 1)} & \multicolumn{1}{c|}{(50, 1)} & Moving feature \\ \hline
\multicolumn{1}{|l|}{\begin{tabular}[c]{@{}l@{}}Concatenate \\ (Moving output)\end{tabular}} & \multicolumn{1}{l|}{($\tau$, 2)} & \multicolumn{1}{c|}{(50, 2)} & \begin{tabular}[c]{@{}l@{}}Moving features from mean of subcarriers over time (50, 1) \\ and previous dense network (50, 1)\end{tabular} \\ \hline\hline
\multicolumn{4}{|c|}{\textbf{Dual Conditional Feature Extraction}} \\ \hline
\multicolumn{1}{|c|}{Network Layer} & \multicolumn{1}{c|}{Symbol} & \multicolumn{1}{c|}{Dimension} & \multicolumn{1}{c|}{Note} \\ \hline
\multicolumn{1}{|l|}{\begin{tabular}[c]{@{}l@{}}Input from previous \\ static output\end{tabular}} & \multicolumn{1}{l|}{($\tau$, $K\times Q + |\boldsymbol{c}^p|$)} & \multicolumn{1}{c|}{(50, 288)} &  \\ \hline
\multicolumn{1}{|l|}{1st GRU (Static)} & \multicolumn{1}{l|}{($\tau$, \#GRU cell)} & \multicolumn{1}{c|}{(50, 32)} & \multirow{5}{*}{\begin{tabular}[c]{@{}l@{}}- Conduct batch normalization for all static/moving GRUs\\ - Recurrent activation function: sigmoid\\ - Output activation function: tanh\end{tabular}} \\ \cline{1-3}
\multicolumn{1}{|l|}{2nd GRU (Static)} & \multicolumn{1}{l|}{($\tau$, \#GRU cell)} & \multicolumn{1}{c|}{(50, 32)} &  \\ \cline{1-3}
\multicolumn{1}{|l|}{\begin{tabular}[c]{@{}l@{}}Input from previous \\ moving output\end{tabular}} & \multicolumn{1}{l|}{($\tau$, 2)} & \multicolumn{1}{c|}{(50, 2)} &  \\ \cline{1-3}
\multicolumn{1}{|l|}{1st GRU (Moving)} & \multicolumn{1}{l|}{($\tau$, \#GRU cell)} & \multicolumn{1}{c|}{(50, 32)} &  \\ \cline{1-3}
\multicolumn{1}{|l|}{2nd GRU (Moving)} & \multicolumn{1}{l|}{(1, \#GRU cell)} & \multicolumn{1}{c|}{(1, 32)} &  \\ \hline\hline
\multicolumn{4}{|c|}{\textbf{Self-Attention}} \\ \hline
\multicolumn{1}{|c|}{Network Layer} & \multicolumn{1}{c|}{Symbol} & \multicolumn{1}{c|}{Dimension} & \multicolumn{1}{c|}{Note} \\ \hline
\multicolumn{1}{|l|}{\begin{tabular}[c]{@{}l@{}}Input from \\ previous block\end{tabular}} & \multicolumn{1}{l|}{($\tau$, \#GRU cell)} & \multicolumn{1}{c|}{(50, 32)} & Input from 2nd GRU (Static) output feature \\ \hline
\multicolumn{1}{|l|}{Dense} & \multicolumn{1}{l|}{($\tau$, \#Output)} & \multicolumn{1}{c|}{(50, 3)} & No activation function \\ \hline
\multicolumn{1}{|l|}{Dense} & \multicolumn{1}{l|}{($\tau$, \#Output)} & \multicolumn{1}{c|}{(50, 1)} & Activation function: sigmoid \\ \hline
\multicolumn{1}{|l|}{Flatten} & \multicolumn{1}{l|}{($\tau$)} & \multicolumn{1}{c|}{(50)} &  \\ \hline
\multicolumn{1}{|l|}{Softmax} & \multicolumn{1}{l|}{($\tau$)} & \multicolumn{1}{c|}{(50)} &  \\ \hline
\multicolumn{1}{|l|}{Multiplication} & \multicolumn{1}{l|}{(1, \#GRU cell)} & \multicolumn{1}{c|}{(1, 32)} & \begin{tabular}[c]{@{}l@{}}Matrix multiplication on input (50, 32) \\ and softmax output (50)\end{tabular} \\ \hline\hline
\multicolumn{4}{|c|}{\textbf{Output}} \\ \hline
\multicolumn{1}{|c|}{Network Layer} & \multicolumn{1}{c|}{Symbol} & \multicolumn{1}{c|}{Dimension} & \multicolumn{1}{c|}{Note} \\ \hline
\multicolumn{1}{|l|}{\begin{tabular}[c]{@{}l@{}}Input from \\ previous block\end{tabular}} & \multicolumn{1}{l|}{(1, \#GRU cell), (1, \#GRU cell)} & \multicolumn{1}{c|}{(1, 32), (1, 32)} & From self-attention output and 2nd GRU (Moving) \\ \hline
\multicolumn{1}{|l|}{Concatenate} & \multicolumn{1}{l|}{(1, \#GRU cell $+$ \#GRU cell)} & \multicolumn{1}{c|}{(1,64)} &  \\ \hline
\multicolumn{1}{|l|}{Flatten} & \multicolumn{1}{l|}{(\#GRU cell $+$ \#GRU cell)} & \multicolumn{1}{c|}{(64)} &  \\ \hline
\multicolumn{1}{|l|}{Dense} & \multicolumn{1}{l|}{(\#Output)} & \multicolumn{1}{c|}{(32)} & Activation function: tanh \\ \hline
\multicolumn{1}{|l|}{Dense (Prediction)} & \multicolumn{1}{l|}{(\#Prediction case)} & \multicolumn{1}{c|}{(4)} & Activation function: softmax \\ \hline
\end{tabular}
\label{NNset}
\end{table*}

\subsubsection{Time Domain Feature Selection}
After receiving $\boldsymbol{G}^p$ from the conditional spatial feature extractor, we implement time domain feature selection using self-attention. Due to the success of attention-based methods in recent years, such as in neural machine translation \cite{21,22}, diverse attention-based applications have emerged. The network architecture of our time domain feature selection is illustrated in Figure \ref{fig:Network}. We aim to train the weight with itself using the conditional spatial feature, denoted as $\boldsymbol{G}^p$. Therefore, we input $\boldsymbol{G}^p$ into a fully connected neural network to convert it into a one-dimensional vector. Then, the output combined vector forms a raw weight vector, which we pass through a softmax function to obtain a weight vector with a total sum of one, denoted as the time weight. Finally, we perform an element-wise product on the received time weight vector to weight $\boldsymbol{G}^p$.

The main idea behind time feature selection is to provide a weighted conditional spatial feature that emphasizes certain time periods with higher weights while ignoring some insignificant time periods to improve performance. Since the conditional spatial feature contains information about the CSI amplitude shape, which varies with time, it includes both varying and non-varying features in a window. We treat these two types of features as distinct aspects and aim to handle them differently. Therefore, we require a network mechanism that can consider these two types of features and assess their importance to execute time selection. For example, when given a window of time features that includes both rapidly varying and non-varying information, the network will automatically prioritize the more important message. To achieve this goal, we use self-attention to emphasize the important weights depending on the time feature obtained after the GRU in order to provide more precise information for classification.

\subsubsection{Feature Mapping and Loss Function}
After obtaining the spatial and moving features from the previous processes, feature mapping concatenates these features and uses them for classification. The concatenated features are then input into a deep neural network to extract appropriate characteristics for the corresponding classes. The prediction output of the network can be attained as
\begin{align}
	\boldsymbol{\hat{o}}^p = \left[ {\hat{o}}^p_1,{\hat{o}}^p_2,{\hat{o}}^p_3,{\hat{o}}^p_4 \right]^{T}
\label{eq:prob}
\end{align}
with elements representing the predicted probability of each case of the $p$-th transmission pair. We compare them with the ground truth labels after one-hot encoding denoted as
\begin{align}
	\boldsymbol{o}^p = \left[ o^p_1,o^p_2,o^p_3,o^p_4 \right]^{T}.
\end{align}
We adopt the cross-entropy to calculate classification accuracy. Consequently, the designed loss function for calculating classification accuracy is leveraged with conventional cross-entropy and $\eqref{eq:con_loss}$ in spatial domain feature extraction as
\begin{align}
	l^p &=-\sum_i{o_i^p \log \hat{o}_i^p}  +   \lambda \cdot l^{p}_{cond} ,
\end{align}
where $i$ is the case index. The significance weight of the loss function $\eqref{eq:con_loss}$ is denoted as $\lambda$. By updating the weights and biases to minimize the loss, we are able to complete our offline model.

\subsection{Voting Scheme}
\label{vote}
In the online phase, as depicted in Fig \ref{fig:blockDiagram}, we perform the same \preproc data preprocess on the received CSI amplitude as in the offline phase. The resulting features are then input into the \netw model which is trained offline. Using the obtained probability for each pair, we construct the probability matrix for all $P$ transmission pairs, which can be expressed as
\begin{align}
	\boldsymbol{\hat{O}} = \left[ \boldsymbol{\hat{o}}^1,\boldsymbol{\hat{o}}^2,\dots,\boldsymbol{\hat{o}}^p,\dots,\boldsymbol{\hat{o}}^P \right] ,
\end{align}
where $\boldsymbol{\hat{o}}^p$ is the probability vector of the four cases of $p$-th transmission pair acquired in $\eqref{eq:prob}$. Since each transmission pair can predict the presence of two adjacent rooms, the probability of the four cases can be combined into a binary case for each room, representing either the absence or presence of people in the room. To achieve this, we perform probability merging, which involves combining the probabilities in a mathematical form of $[empty, presence]$. For the rooms deployed with RX, having their probabilities of each case, the merged probability of the $p$-th transmission pair is obtained as
\begin{align}
	\boldsymbol{v}^p_{RX}= \left[ {\hat{o}}^p_1+{\hat{o}}^p_2,{\hat{o}}^p_3+{\hat{o}}^p_4 \right] ,
\end{align}
where cases 1 and 2 are the states with no people in RX room and cases 3 and 4 represent the state with people in RX room. For the TX room, we can similarly acquire the merged probability of the $p$-th transmission pair given by
\begin{align}
	\boldsymbol{v}^p_{TX}= \left[{\hat{o}}^p_1+{\hat{o}}^p_3,{\hat{o}}^p_2+{\hat{o}}^p_4 \right] .
\end{align}
It is important to note that the empty cases for the TX room are cases 1 and 3, while cases 2 and 4 represent the presence of people in the room. Note that the probability to be added is different from those for RX room. As receivers share only one transmitter, we denote that all the probabilities of TX room from all $P$ transmission pairs as $[\boldsymbol{v}^1_{TX},\boldsymbol{v}^2_{TX},\dots,\boldsymbol{v}^p_{TX},\dots,\boldsymbol{v}^P_{TX}]$. We then perform multi-room voting by computing the average probability for each transmission pair, which can be expressed as
\begin{align}
	\boldsymbol{v}_{TX} = \frac{1}{P} \sum^P_{p=1}{\boldsymbol{v}^p_{TX}} .
\end{align}
For a total of $R$ rooms, we utilize the probability vector of each pair in the RX rooms and execute a decision by choosing the higher probability as the final prediction for the $R-1$ rooms. Similarly, for the $R$-th room deployed with TX, we determine by selecting the case with higher probability in $\boldsymbol{v}_{TX}$ to obtain the prediction for room $R$. In this way, we can obtain the predictions for all $R$ rooms. By adopting the proposed voting mechanism, the classification performance of the room deployed with TX can be effectively enhanced with stabilized prediction performance. In Table \ref{NNset}, we have listed all the parameter settings and corresponding dimension of the deep neural network of the proposed TCD-FERN scheme. Note that the optimal dimension of window size and condition input size will be discussed in the following section.

\section{Methodology} \label{CHP_ME}

\subsection{Experimental Settings}

\begin{table}[!t]
\small
\centering
\caption{Parameters of WiFi Sensing System}
\begin{tabular}{ll}
\hline
Parameters of system  	& Value            \\ \hline \hline
Carrier frequency   	& 2.447 GHz        \\
Channel bandwidth     	& 20 MHz           \\
Number of subcarrier 	& 56               \\
Number of antenna pair 	& 4                \\
Number of RX  			& \{1, 2\}         \\
Number of room  		& \{2, 3\}         \\
Number of WiFi AP    	& \{2, 3\}         \\
Data collection rate   	& 10 packets/sec   \\
Sample number    		& \{20000, 16000\}  \\
Significance weight $\lambda$                                         & 0.5                  \\ \hline
\end{tabular}
\label{systemPar}
\end{table}

\begin{figure*}[t]
\centering
\subfigure[]{ \includegraphics[width=3.3in]{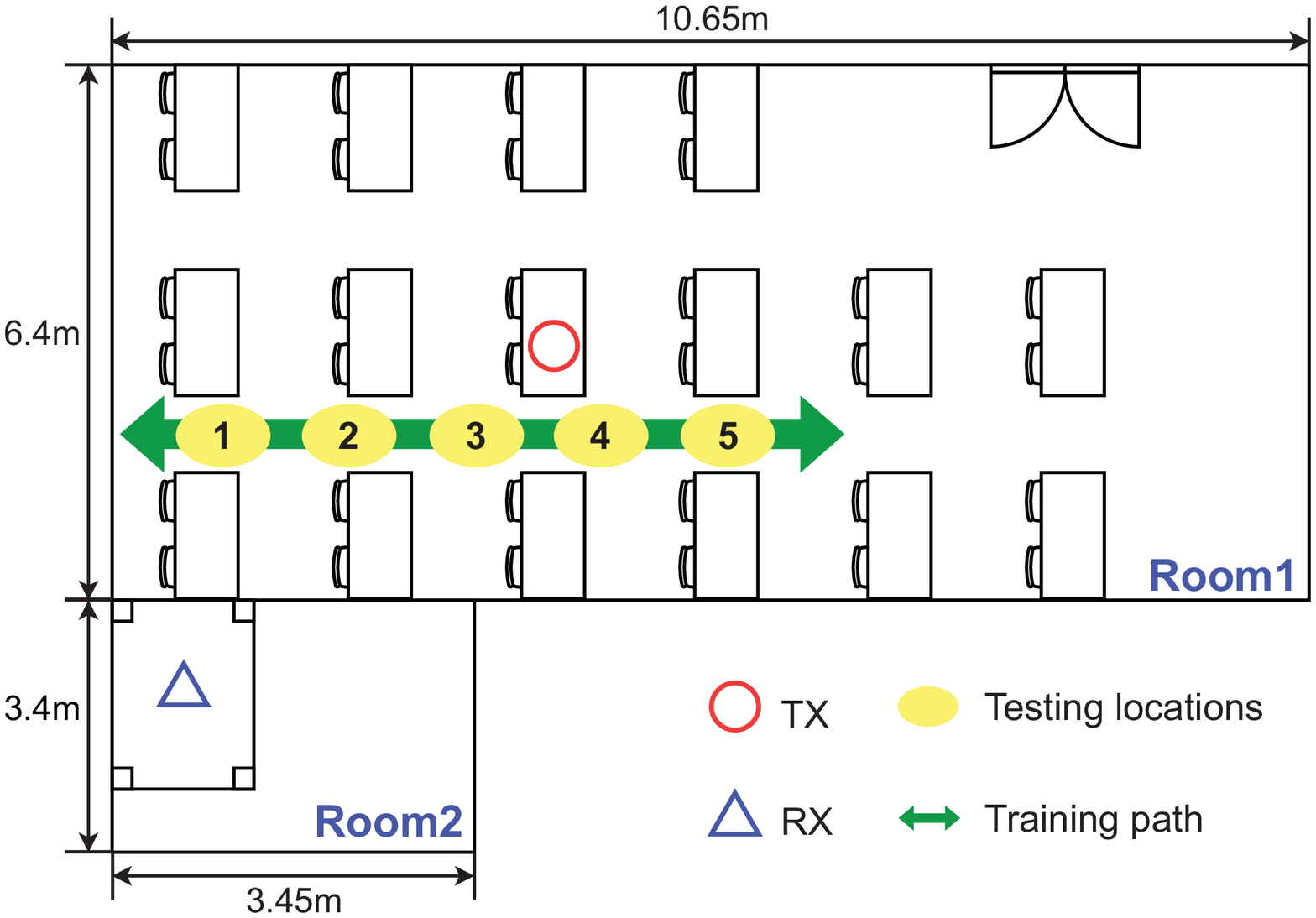} \label{fig:scen1} }
\quad
\subfigure[]{ \includegraphics[width=3.3in]{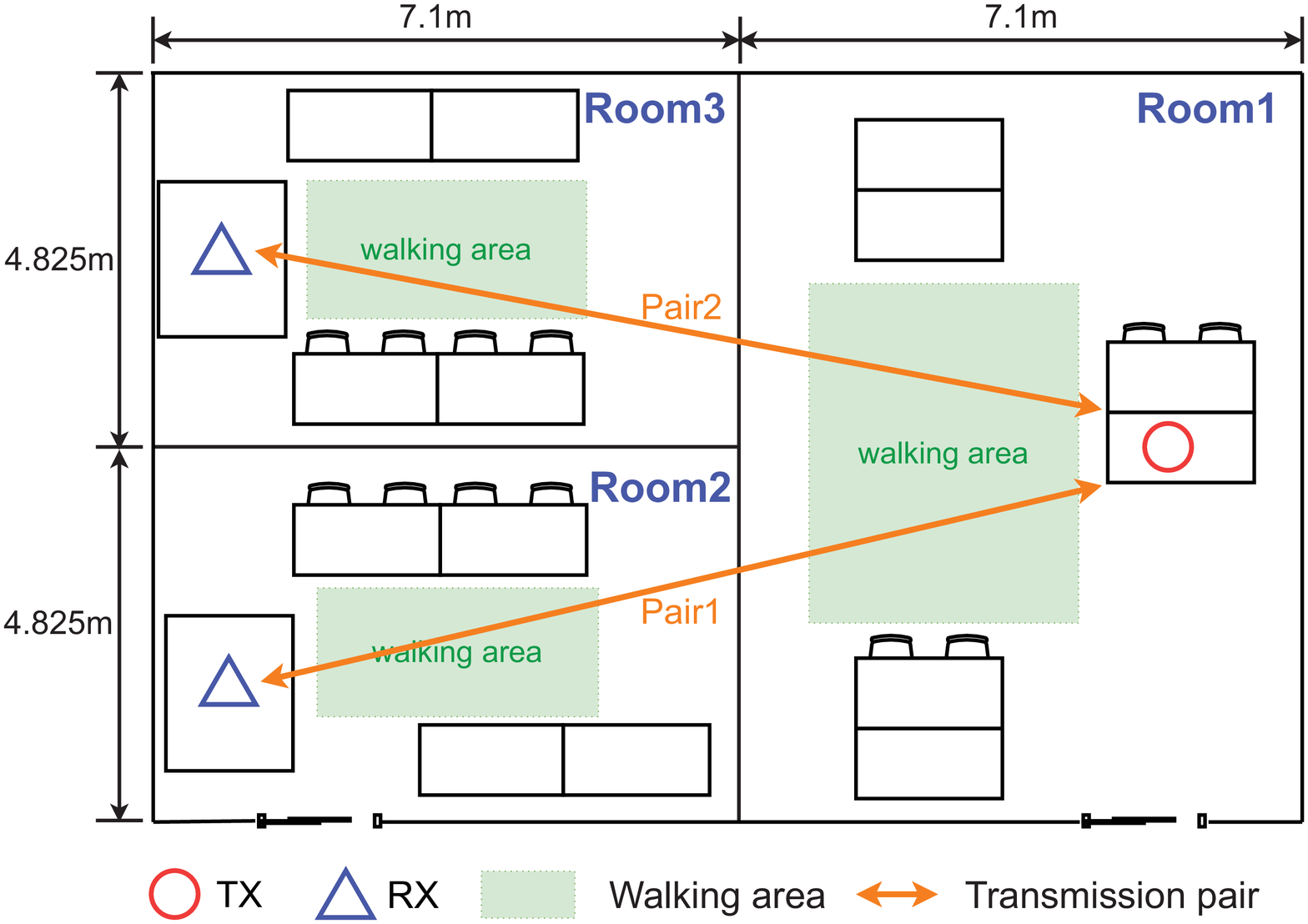}\label{fig:scen2}	}
\caption{The experimental scenarios and setting in (a) a laboratory and (b) a conference room.}
\label{ExpScen}
\end{figure*}

\begin{figure*}[t]
\centering
\subfigure[]{\includegraphics[width=0.47\linewidth]{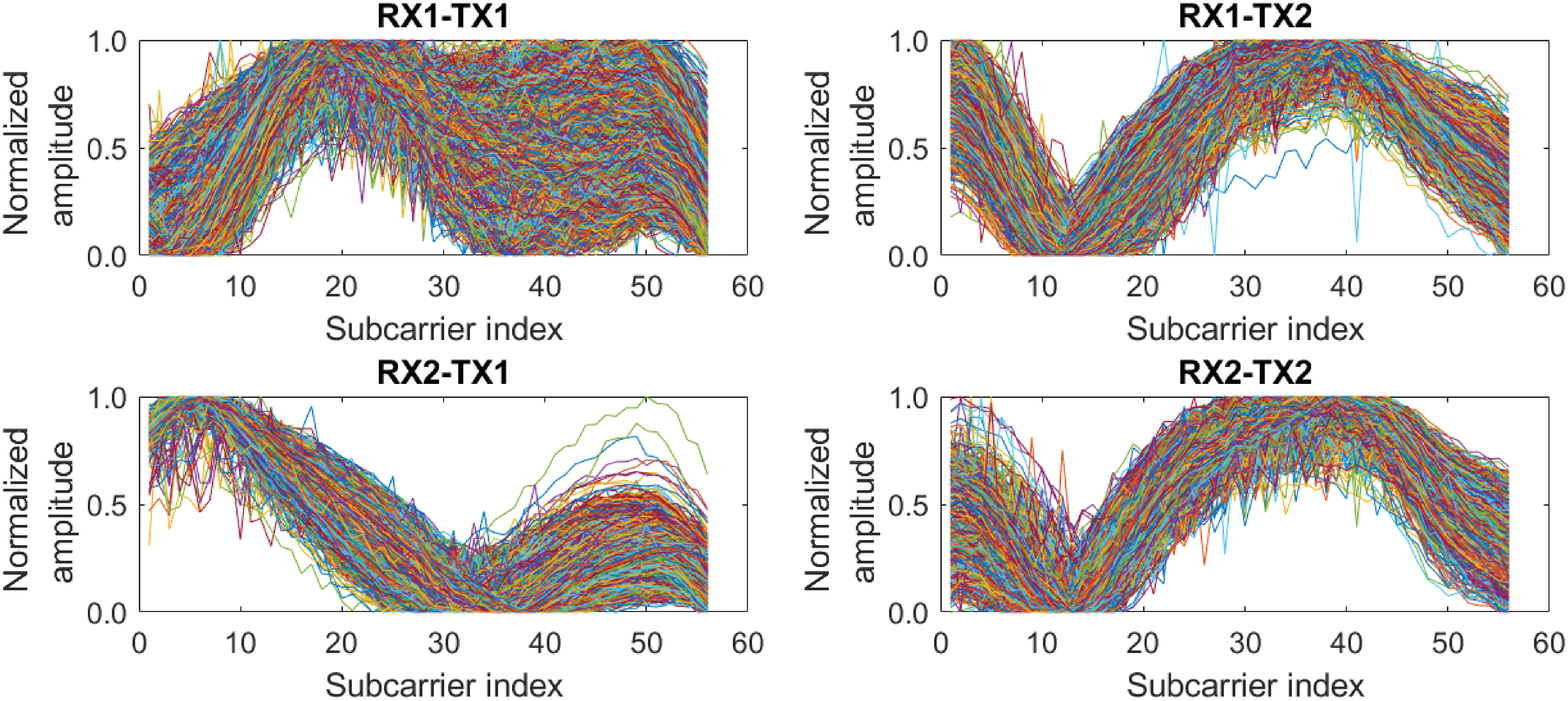}\label{fig:TX_side}	}
\quad
\subfigure[]{ \includegraphics[width=0.47\linewidth]{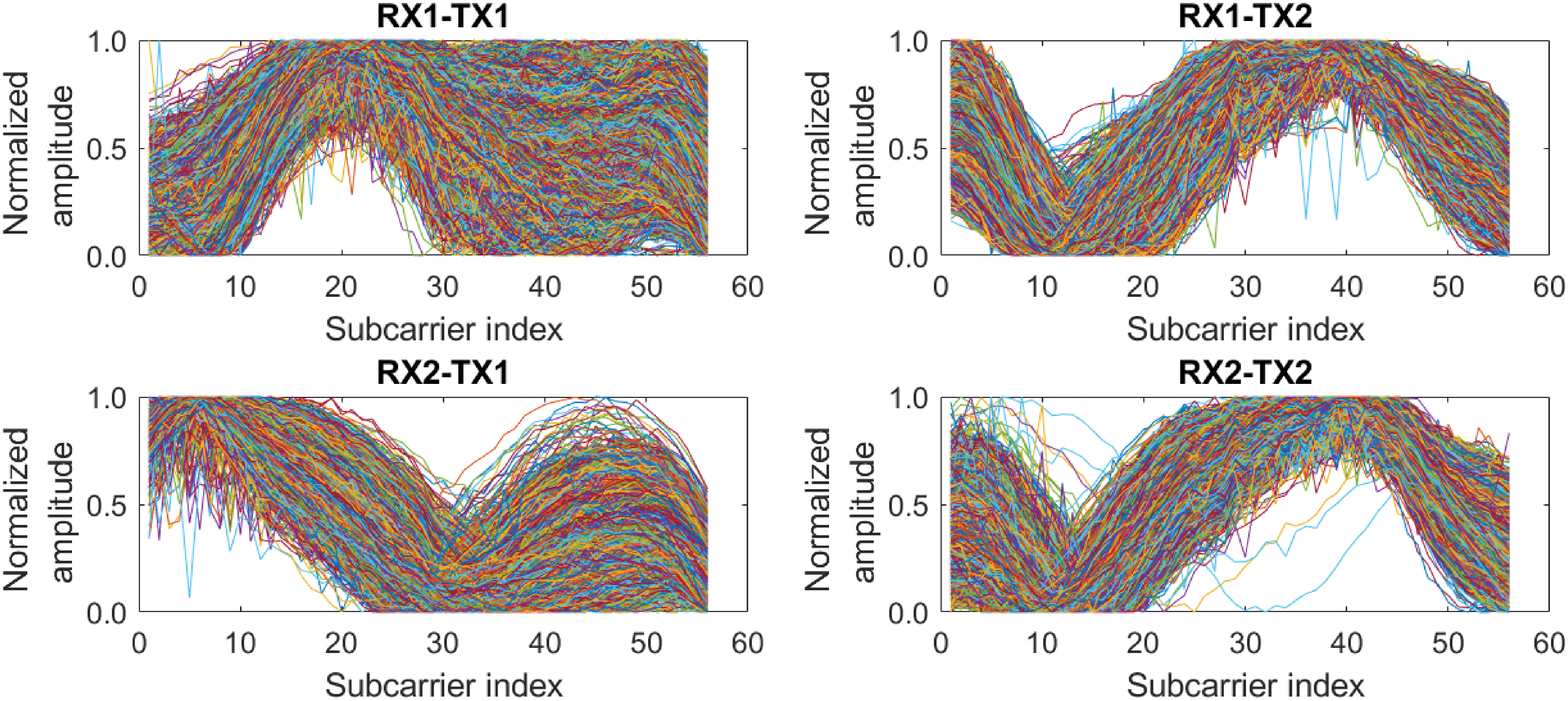}\label{fig:RX_side}	}
\quad
\subfigure[]{\includegraphics[width=0.47\linewidth]{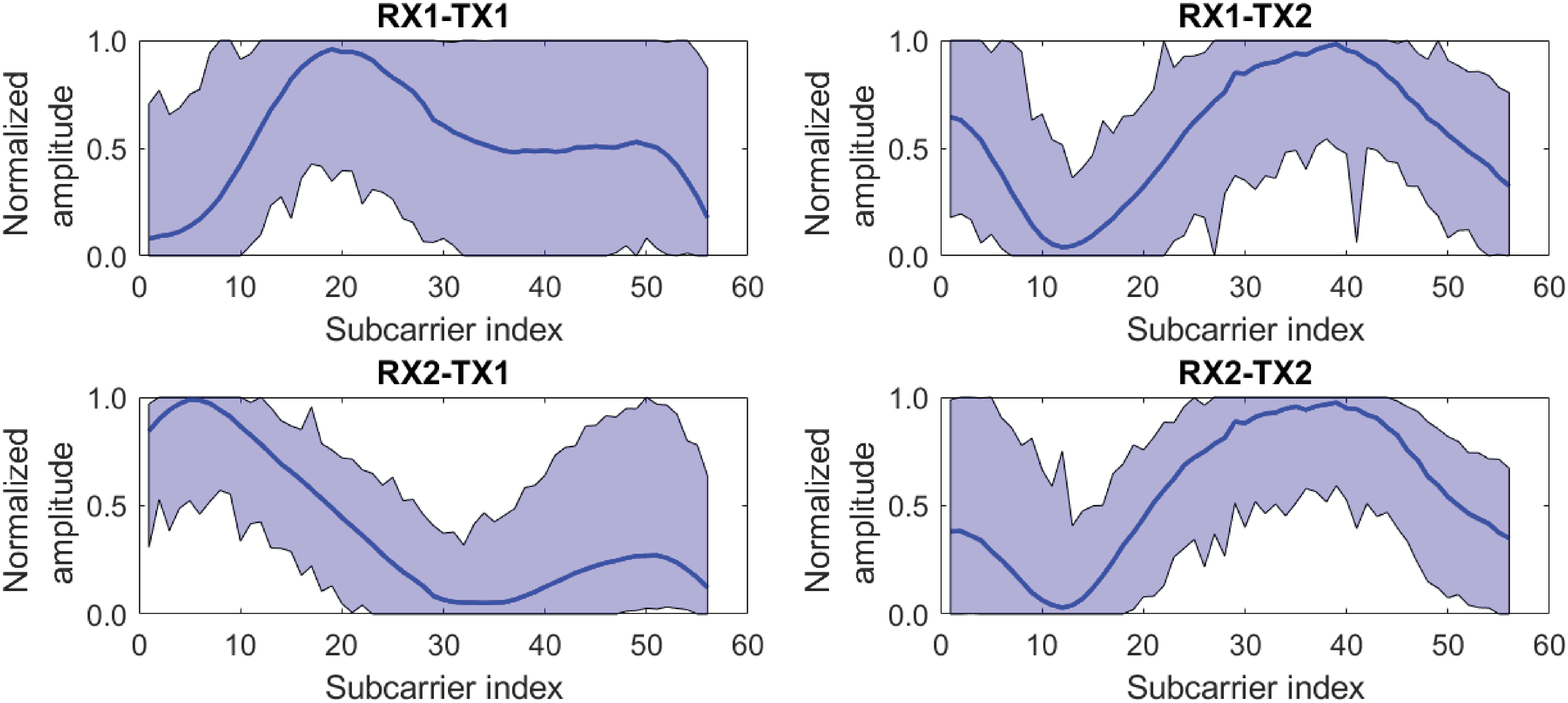}\label{fig:TX_side1}	}
\quad
\subfigure[]{\includegraphics[width=0.47\linewidth]{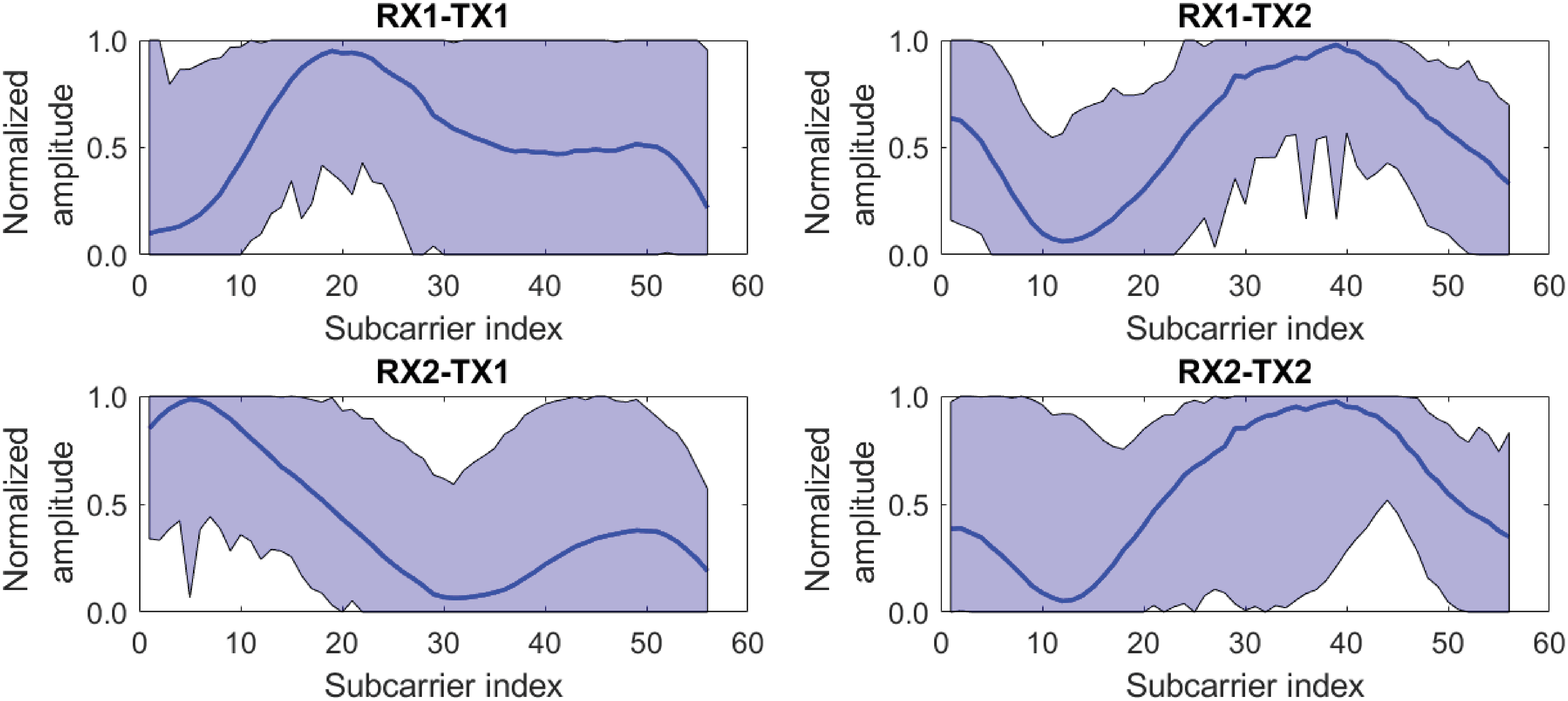}\label{fig:RX_side1}	}
\caption{Normalized CSI amplitude versus subcarrier index for human presence in the room deployed with either (a) transmitter or (b) receiver, with their respective boundary plots in (c) and (d). Note that the solid line indicates the average of CSI amplitude, whilst the colored area represents the boundary of CSI amplitude.}
\label{TXvsRX}
\end{figure*}

In our experiments, we employ TPLink-WDR4300 wireless routers serving as the transmitter and multiple receivers in our star network. The routers are operated based on IEEE 802.11n standard and transmit CSI data in the central carrier frequency of 2.447 GHz with a carrier bandwidth of 20 MHz. Each router is equipped with two antennas, resulting in four antenna pairs in a transmission pair. Each antenna pair contains 56 subcarriers. We conduct experiments in two different scenarios as shown in \fig \ref{ExpScen}. In both experiments, we collect data at a sampling rate of 10 Hz\textsuperscript{\ref{note1}}\footnotetext[1]{As limited by the hardware, we can select a maximum sampling rate of 10 Hz with an AP router of TPLink-WDR4300 (802.11n). Of course, if the state-of-the-art WiFi devices are available, such as WiFi-6/7 (802.11ax/be) \cite{conf} with higher processing speed and wider bandwidth, we can collect higher-resolution dataset with corresponding higher detection accuracy performance. \label{note1}}, which includes training data for the offline phase and testing data for the online phase. Note that we have carried out cross-validation by splitting $20\%$ training data as our validation data.

\begin{itemize}
	\item The first experiment, shown in \fig \ref{fig:scen1}, is conducted in two neighboring labs to create a two-room scenario, including a larger lab with a size of $10.65m \times 6.4m$ and a smaller lab with a size of $3.45m \times 3.4m$. The setup includes a TX in room 1 and an RX in room 2, with both routers placed on tables. Testers walk back and forth along the green arrow path in \fig \ref{fig:scen1} to collect training data, while testing data is gathered from the five yellow elliptical areas, where people are either walking or standing still in each small elliptical area. The five testing locations represent three states: LoS path blocking at location 3, NLoS with abundant multipath at locations 1 and 2, and NLoS with scarce signal reflection at locations 4 and 5. We have collected 2000 data for each case (both rooms empty, presence in either TX or RX room, and presence in both rooms), resulting in a total of 8000 data for the four cases as our training data. We also collect 1000 data per case for each testing location, which gives a total of 4000 data for each location, and a total of 20000 data for five locations.

	\item The second experiment is conducted in a conference room to test a multi-room scenario as shown in \fig \ref{fig:scen2}. The conference room had a size of $14.2m \times 9.65m$ and is divided into three rooms. Room 1 is deployed with a transmitter, while rooms 2 and 3 are equipped with receivers all placed on tables. Since there are three rooms in this scenario with one transmitter and two receivers, it offers two transmission pairs. Testers walk arbitrarily in the green area shown in \fig \ref{fig:scen2}, halting at random spots frequently. In addition, there are eight presence cases for the three-room scenario. We have collected 5000 data for each case, resulting in a total of 40000 training data, and 2000 data per case for testing, resulting in a total of 16000 testing data.

\end{itemize}

\subsection{Observations of Human Effects on Wireless Signals}

Since human presence can cause fluctuations in WiFi signals due to multipath propagation, we conducted preliminary experiments to investigate the effects of human presence on wireless signals under different conditions. As described in Section \ref{CHP_PRE}, our system architecture consists of a TX and multiple RXs. However, since each transmission pair is independent when analyzing CSI received from each RX, we employed CSI from one of the transmission pairs of two adjacent rooms as a representation for our observations. Based on these CSI data, we obtain the following characteristics related to the effects of human presence at the TX and RX sides, as well as the effects of LoS path blocking and non-blocking.

\subsubsection{Effects of Human Presence at TX side and RX side}
\label{TXRXside}

As our system utilizes TTW signals to detect two rooms with only a pair of TX and RX, we can observe the differences in signals when there is human presence at the TX or RX side. An example of the CSI amplitude versus subcarrier index for these two situations is shown in Fig. \ref{TXvsRX}. Figs. \ref{fig:TX_side} and \ref{fig:RX_side} show the all normalized CSI amplitudes, whereas Figs. \ref{fig:TX_side1} and \ref{fig:RX_side1} reveal the averaged CSI amplitude (solid line) and the corresponding boundary (colored area). From Figs.  \ref{fig:TX_side} and \ref{fig:TX_side1}, we can be observe a smaller variance of CSI amplitude for different subcarriers when a person is walking in the room with the TX compared to when a person is walking in the room with the RX in Figs. \ref{fig:RX_side} and \ref{fig:RX_side1}. This phenomenon is more obvious within the pairs of RX2-TX1 and RX2-TX2. This could be due to the wall between the two rooms. The signal sent from the transmitter passes through the wall after being influenced by human presence in the room with the TX, causing the feature generated by humans to become relatively small compared to the effect from the wall. On the contrary, when the signal is influenced after passing through the wall, the receiver would receive the signal right after it is affected by human, making the fluctuations more significant. Moreover, the propagation distance of the signals affected in the TX room is longer, which causes the signals to encounter more blocking and reflections, resulting in more severe attenuation of the feature in the signal. Therefore, it is important to focus on the design of capturing important feature from the TX room and improving performance of human presence in the TX room.

\subsubsection{Effects of LoS path Blocking and Non-Blocking}
\label{threeStateBlocking}

\begin{figure}[t]
\centering
\subfigure[]
{\includegraphics[width=2.5in]{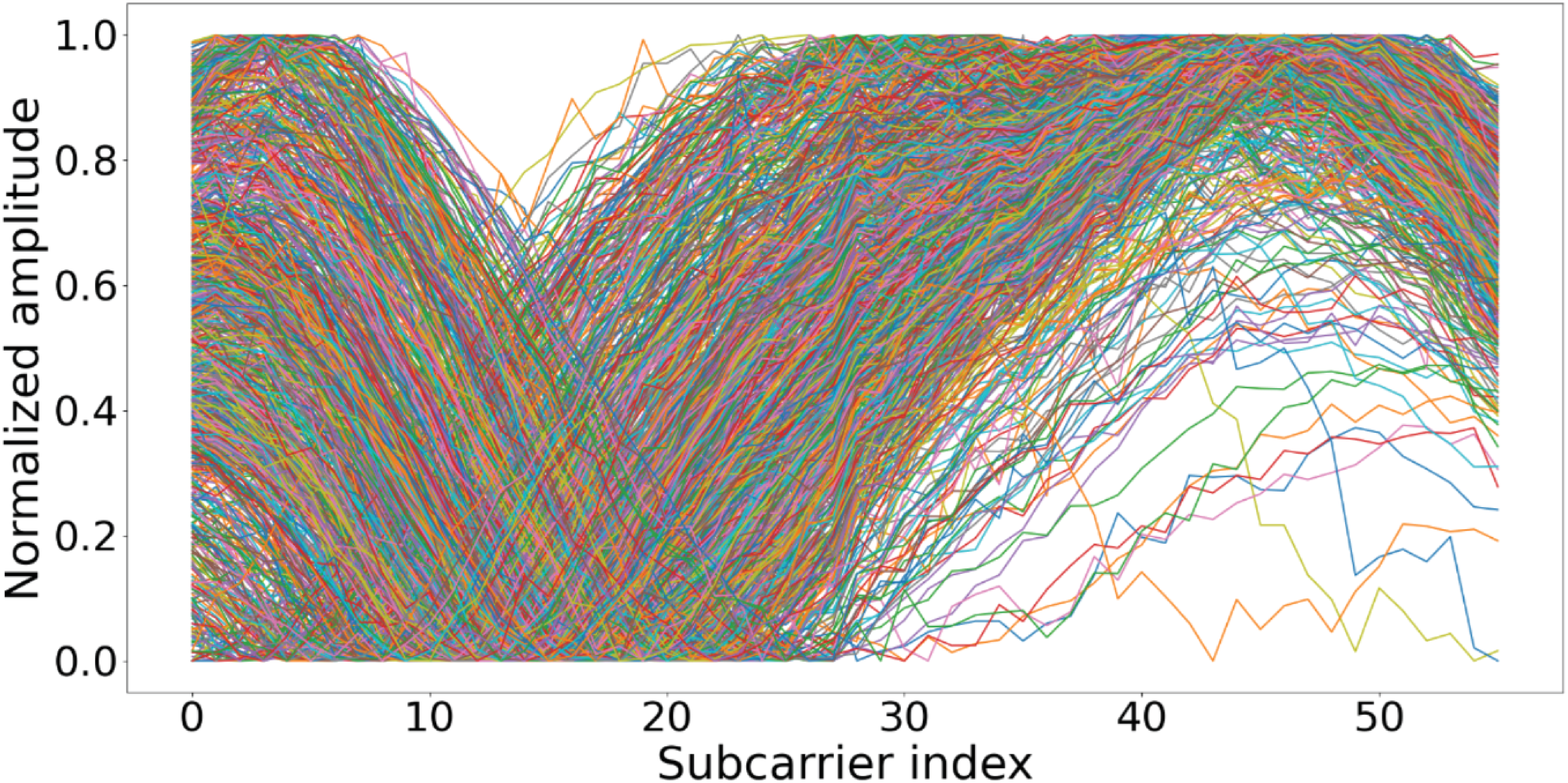} \label{fig:los}	}
\quad
\subfigure[]{\includegraphics[width=2.5in]{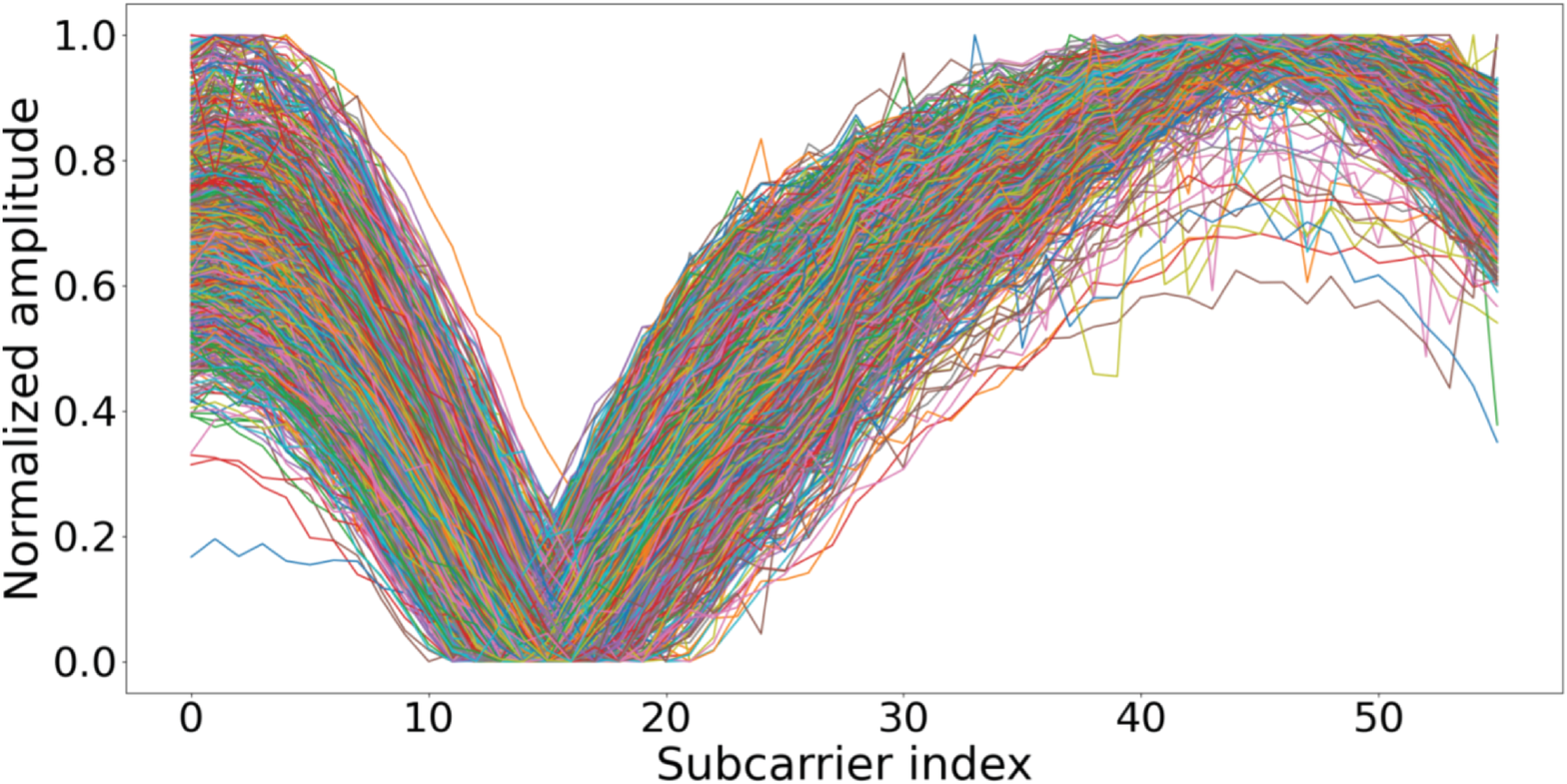}\label{fig:wall}}
\quad
\subfigure[]{\includegraphics[width=2.5in]{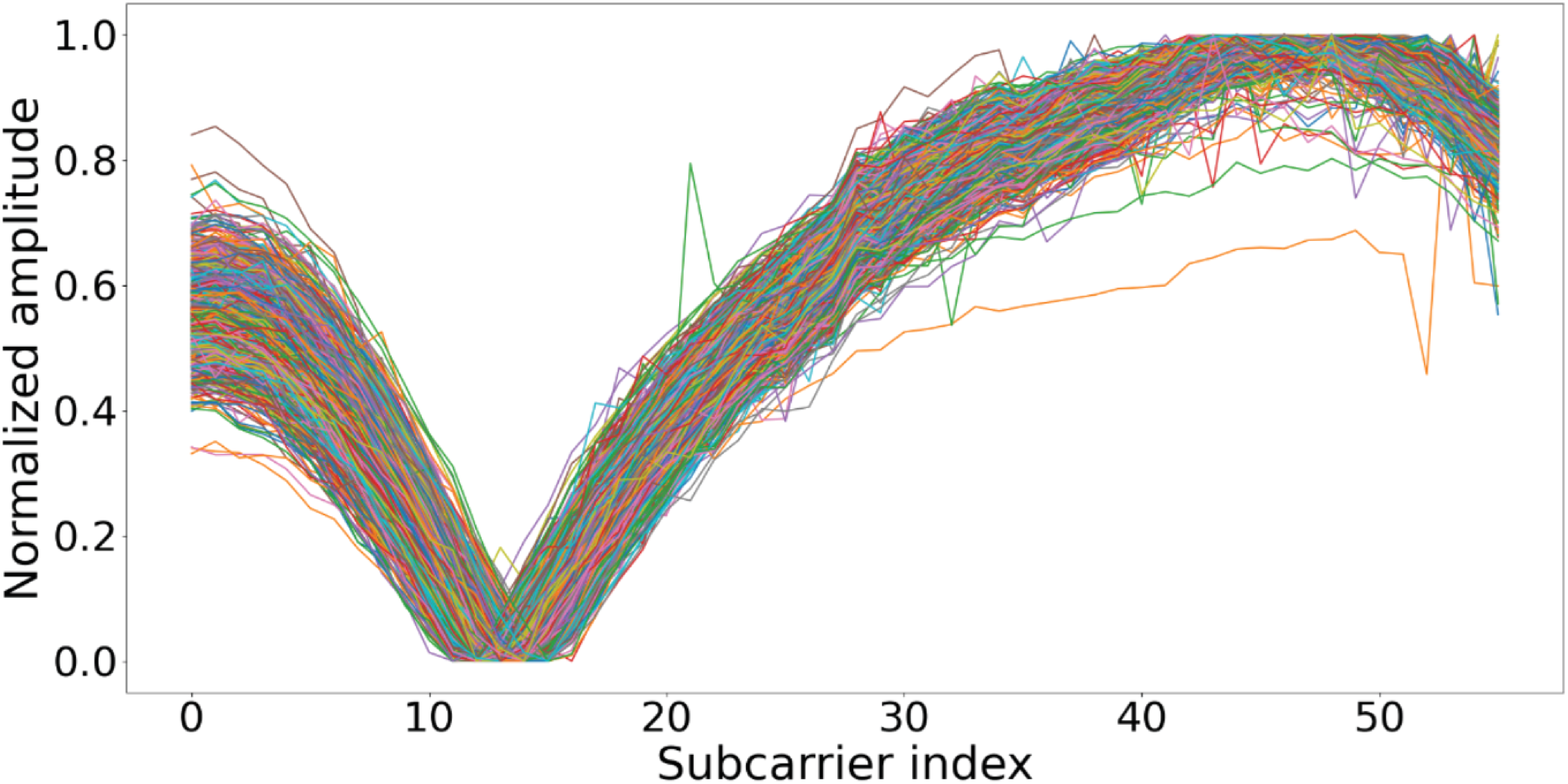} \label{fig:nlos}}
\caption{Normalized CSI amplitude of human presence at (a) LoS, (b) NLoS in reflective space, and (c) NLoS in open space.}
\label{LoSvsNLoS}
\end{figure}

We have also conducted experiments to investigate the effects of human presence on CSI amplitude in different environmental situations, where people are located in either LoS or NLoS positions. LoS path blocking occurs when a person is in the direct path, while LoS non-blocking happens when the person is not blocking the direct path. For the LoS non-blocking state, we further divide it into two classes: environments with rich multipath such as narrow rooms with walls, and open spaces with no signal reflectors. \fig \ref{fig:los} displays the CSI amplitude of human presence at LoS, where the variance of the subcarriers of different packets is the largest among the three subplots. Conversely, the amplitude variances of human presence at NLoS in \fig \ref{fig:wall} and \fig \ref{fig:nlos} are relatively smaller. This phenomenon can be explained by the direct impact of the person blocking the LoS signal path propagating to the RX, making the received CSI amplitude sensitive to changes caused by humans. For the NLoS case, we can observe that the variance of the CSI amplitude of human presence in an open space in \fig \ref{fig:nlos} is comparably smaller than that in a reflective space in \fig \ref{fig:wall}. This is because reflective spaces possess more severe multipath compared to open spaces. In open spaces, with a lack of reflection, the RX receives fewer signals containing features generated by human presence, resulting in little variation between different packets. This makes it difficult to distinguish the human presence at NLoS in open space from the case of empty room.

\begin{figure}[t]
\centering
\subfigure[]{ \includegraphics[width=0.96\linewidth]{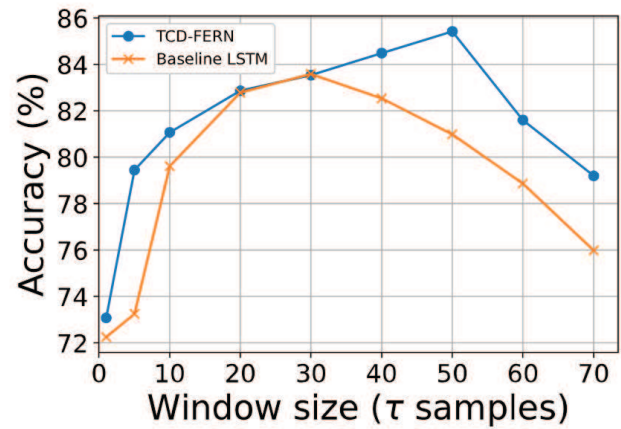}\label{fig:windowsize}}
\quad
\subfigure[]{ \includegraphics[width=0.96\linewidth]{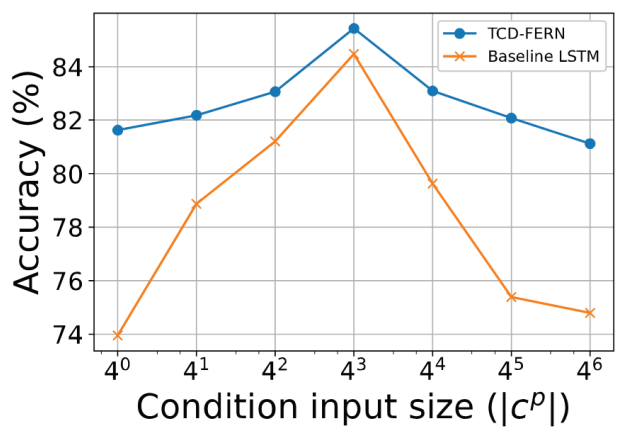} \label{fig:condsize}}
\caption{Average accuracy curves versus (a) different window sizes $\tau$ and (b) different condition input sizes $|\boldsymbol{c}^p|$.}
\label{curve}
\end{figure}

\section{Performance Evaluation} \label{CHP_ER}

\subsection{Impact of Designed Parameters}
In our \netw system, two parameters need to be determined at the beginning of our experiment, i.e., the window size $\tau$ and the size of the condition input. In this subsection, we adopt the datasets of the first experiment with two labs illustrated in \fig \ref{fig:scen1} to evaluate the performance of different parameter sizes. The accuracies presented in this subsection are the averaged accuracies of five testing location datasets.

\subsubsection{Evaluation of Different Window Sizes}
The window size $\tau$ determines the duration of time for which data is imported into our system. It represents the amount of time required by the network to judge the current situation of human presence, i.e., the amount of information required to provide accurate prediction. We apply window sizes ranging from 1 to 70 with an interval of 10 to observe the effect of different window sizes. The average accuracy curve obtained is plotted in \fig \ref{fig:windowsize}. As the window size increases, the accuracy rises due to the increased feature information. However, the prediction accuracy decreases when we provide too much information with a longer window size. This is because the excessive patterns for each prediction may cause overfitting and confusion. Based on the results, we set the window size for the following experiments as $\tau = 50$. To elaborate a little further, we have compared the proposed GRU-based architecture to that of LSTM. Although LSTM can tackle large sequence of data; however, human presence detection usually requires a short time-window size to prevent compellingly high latency. Benefited by the moderate training in both small- and large-size of sequence, proposed TCD-FERN with GRU outperforms the baseline LSTM with an accuracy difference of around $2\%$.

\subsubsection{Evaluation of Different Condition Sizes}
The condition input size determines the size of spatial domain features $|\boldsymbol{c}^p|$ in $\eqref{condii}$ adopted as the input for dual conditional feature extraction. This factor represents the amount of information used as a condition, which provides a certain level of traction in GRU feature extraction. We have tested condition input sizes ranging from $1$ to $4096$. From \fig \ref{fig:condsize}, we observe that the highest accuracy occurs at a condition input size of $64$, and lower accuracy takes place with either smaller or larger condition input sizes. When the condition input size is smaller than that of the major input, it contributes relatively little information, making important features being ignored and performing a worse training result. Conversely, with a larger condition input size, it may dominate the training. Therefore, we consider the condition input size as $64$ in the following evaluations. Furthermore, as explained previously, the proposed TCD-FERN with moderate training achieves higher accuracy than that of the baseline LSTM, with an accuracy improvement of around $2\%$ to $8\%$.

\begin{figure}[!t]
\centering
\subfigure[]{\includegraphics[width=.98\linewidth]{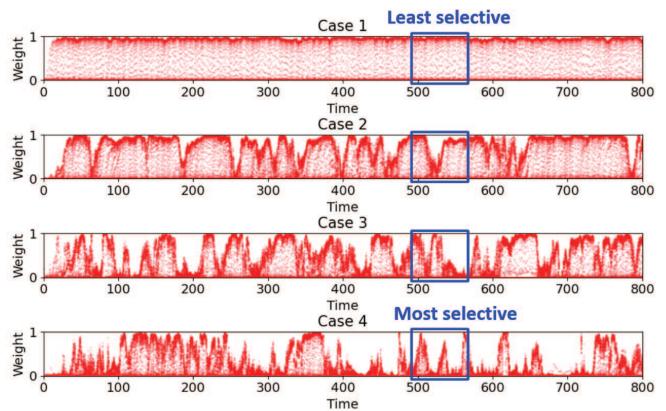}\label{fig:timeSelect1}}
\quad
\subfigure[]{\includegraphics[width=.98\linewidth]{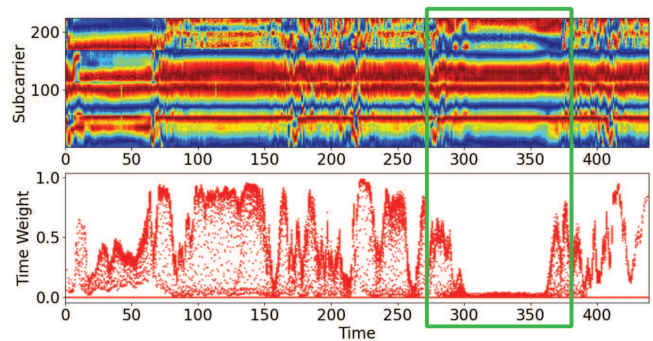} \label{fig:timeSelect2}}
\caption{Analysis of time weight. (a) Comparison of the obtained time weight between the four human presence cases. The blue boxes show the obtained time weight of a window, where case 1 has the least selective time weight and case 4 has the most selective time weight of case 4. (b) Relationship between the CSI amplitude and the obtained time weight. The green box points out a perceptible interval where smaller weight is assigned for smaller fluctuation of CSI amplitude and vice versa.}
\label{timeSelection}
\end{figure}

\subsection{Impact of Time Selection}

We have evaluated the effectiveness of the time weight trained in self-attention and examined the characteristics selected by our model after training. The data used for the analysis is obtained from the first experiment scenario in Fig. \ref{fig:scen1}. We analyze the time weight of different cases and the relationship between CSI amplitude and time weight as shown in \fig \ref{timeSelection}. \fig \ref{fig:timeSelect1} displays the training time weight of the four cases (cases 1 to 4, from top to bottom) and marks the time weights of a window for each case with blue frames. We can observe that case 1 has the least selective time weight, with similar magnitude of weight in a window. On the other hand, case 4 has a highly selective time weight, with significant fluctuation, which is reasonable since case 4 represents human presence in both rooms, leading to significant CSI amplitude variation in time. Hence, it is necessary to select representative time features from numerous time features. Furthermore, we observe that the layout of TX and RX also affects the outcome of the time weight. Although both cases 2 and 3 represent human presence in only one room, the time weight of case 3 is more selective than that of case 2 due to the more prominent features from the RX room.

Furthermore, we aim to gain insights into the learned features of our model. To achieve this, we analyze the relationship between the CSI amplitude and the obtained time weight. We use the data from the first scenario and plot the CSI amplitude segment of case 1 and the corresponding time weight in \fig \ref{fig:timeSelect2}. In the figure, we highlight a noticeable interval with a green frame. It is evident from the green frame that the model assigns a lower time weight when there is almost no fluctuation in CSI amplitude. On the other hand, the model assigns higher weight to the edges of the green frame where there is variation in the corresponding CSI amplitude. Thus, we can infer that the fluctuation in the CSI amplitude is a critical feature for accurate classification.

\begin{figure}[t]
\centering
\includegraphics[width=1\linewidth]{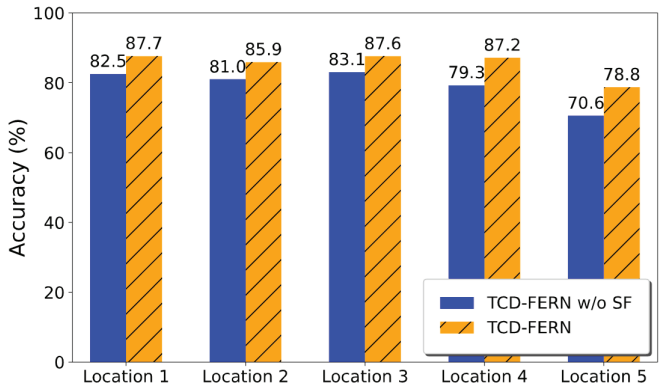}
\caption{Performance comparison of the detection accuracy for five locations between \netw with and without subcarrier fusion (SF), denoted as TCD-FERN and TCD-FERN w/o SF, respectively.}
\label{fig:subFusion}
\end{figure}

\begin{figure}[t]
\centering
\includegraphics[width=1\linewidth]{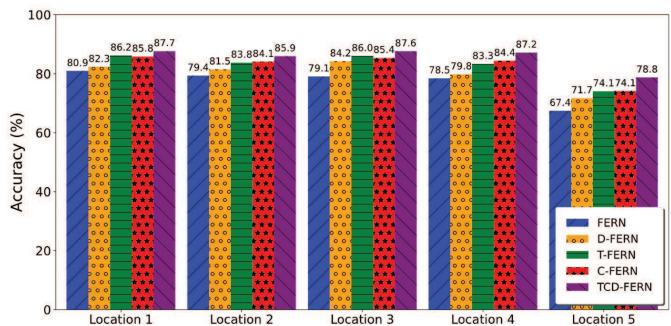}
\caption{Accuracy comparison of \netw with different proposed architecture combination for five testing locations. Note that different colored bars from left to right denote the architecture with only a single GRU (FERN), with dual GRU (D-FERN), adding time selection (T-FERN), adding condition (C-FERN) and proposed architecture (TCD-FERN).}
\label{fig:difArc}
\end{figure}

\begin{figure*}[t]
\centering
\subfigure[]{\includegraphics[width=2.2in]{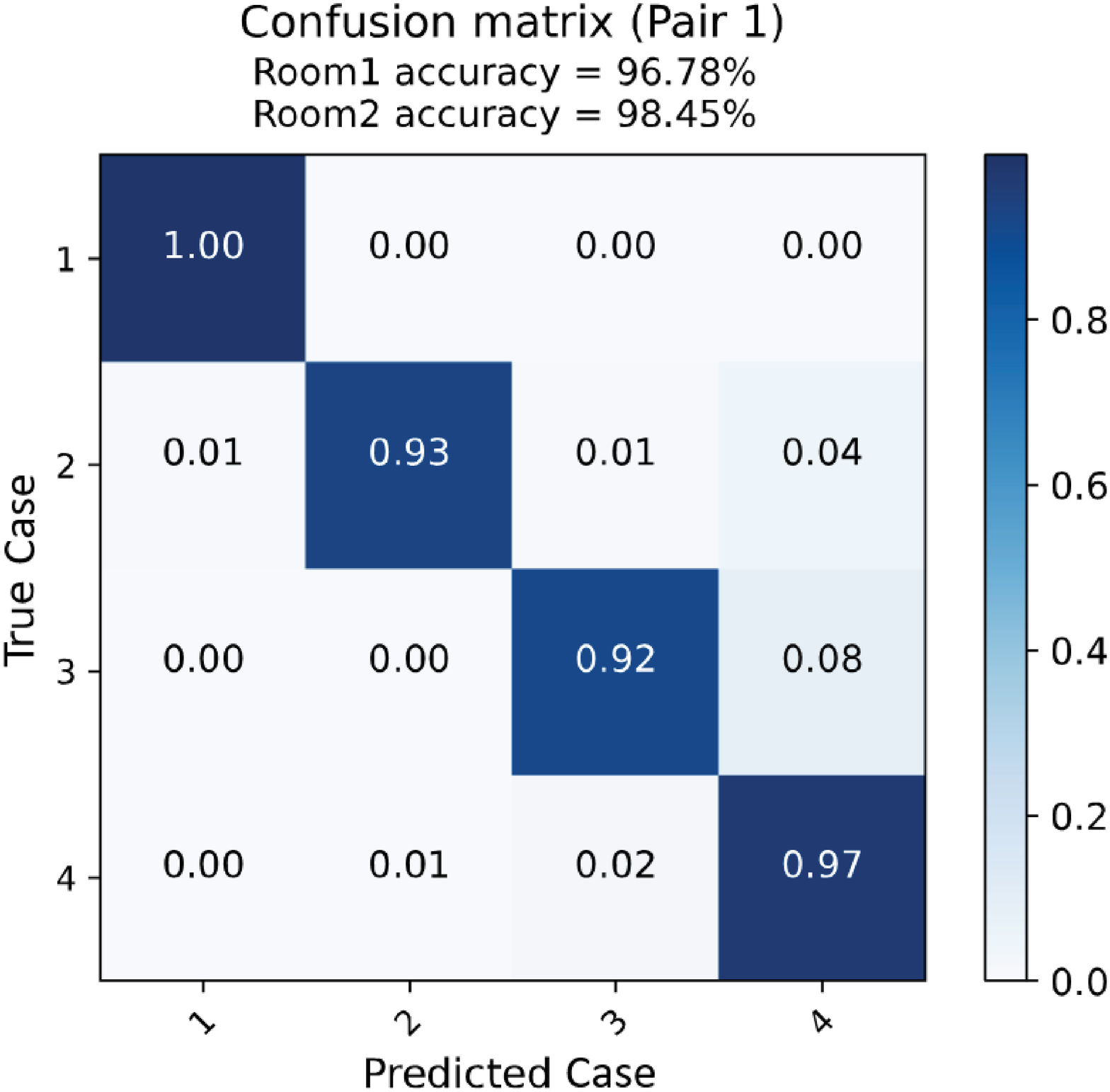} \label{fig:confusion1}}
\quad
\subfigure[]{\includegraphics[width=2.2in]{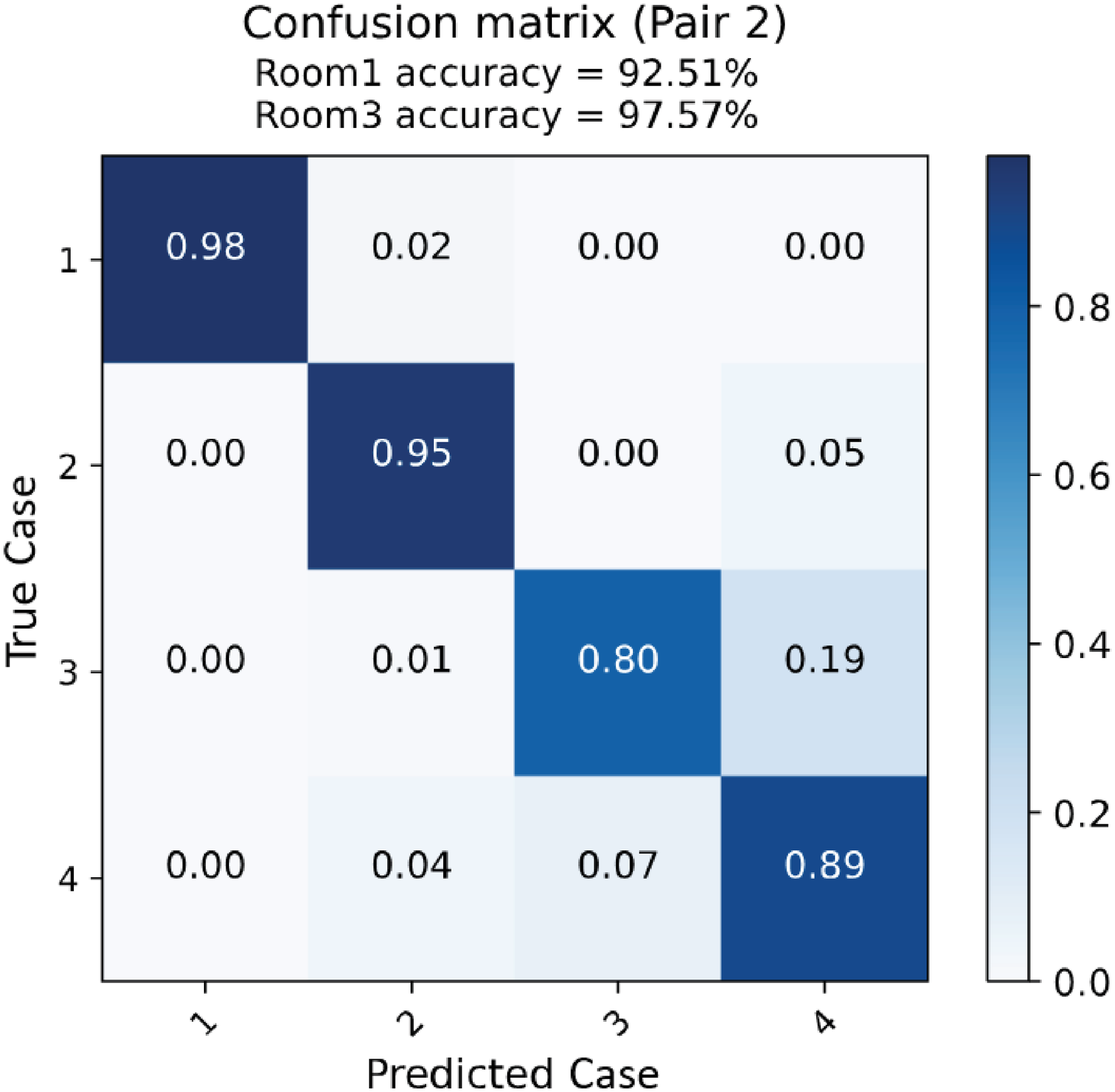} \label{fig:confusion2}}
\quad
\subfigure[]{\includegraphics[width=2.2in]{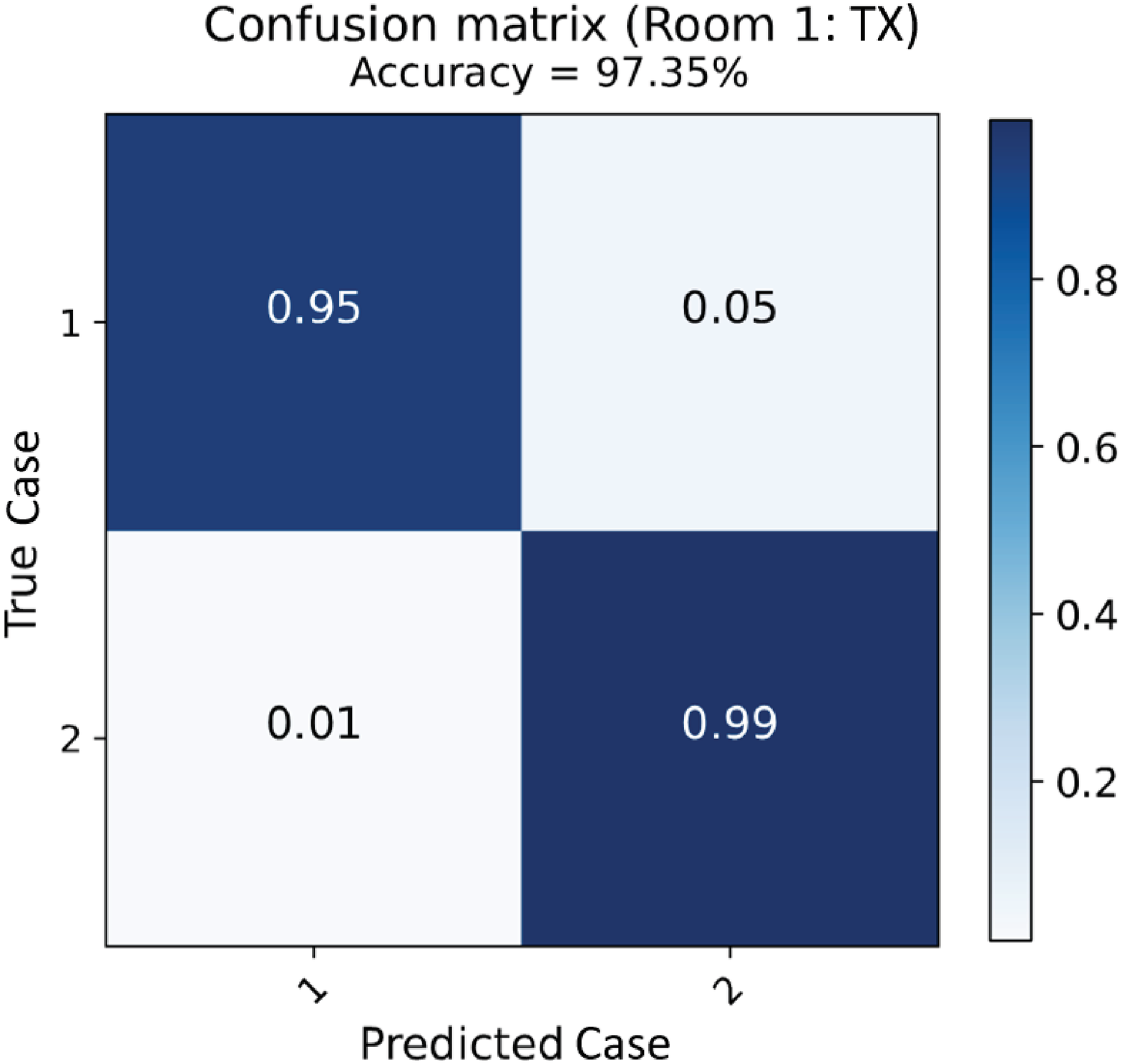} \label{fig:confusionRoom1}}
\caption{Confusion matrix of the two transmission pairs for the four cases in the conference room scenario. (a) Transmission pair 1 with rooms 1 (TX) and 2 (RX): The probabilities of the four cases for rooms 1 and 2 have the accuracy of 96.78\% and 98.45\%, respectively. (b) Transmission pair 2 with rooms 1 (TX) and 3 (RX): The probabilities of the four cases for rooms 1 and 3 have the accuracy of 92.51\% and 97.57\%, respectively. (c) Confusion matrix of room 1 (TX) after voting scheme from transmission pairs 1 and 2. Note that case 1 here indicates an empty room, whereas case 2 represents human presence. The final prediction accuracy for room 1 is 97.35\%.}
\label{fig:pairConfusion}
\end{figure*}

\subsection{Performance of Proposed \netw}

In this subsection, we evaluate the ablation studies of the proposed \netw scheme. As previously mentioned, subcarrier fusion can assist to eliminate confusion caused by abundant multipath and reduce computational complexity. To evaluate the efficiency of subcarrier fusion, we utilize the datasets from the first scenario and analyze the adoption of subcarrier fusion for the five testing locations. \fig \ref{fig:subFusion} shows the detection accuracy of \netw with and without subcarrier fusion. The figure indicates that subcarrier fusion has resulted in improved accuracy across all five locations. This improvement is observed not only in LoS conditions but also in NLoS scenarios with varying degrees of reflection. These results highlight the effectiveness of subcarrier fusion in enhancing performance of the proposed TCD-FERN scheme.

Moreover, we evaluate the performance of the three main architectures in our proposed \netw model, which include the dual GRU network, time selection, and condition input extraction. We employ the datasets of the first scenario for the evaluation. \fig \ref{fig:difArc} shows the detection accuracies of the five testing locations for five different architectures. The first architecture, labeled as FERN, is a commonly adopted network with only a single GRU for training, which serves as our baseline scheme for performance comparison. The second architecture, named D-FERN, represents the model with dual GRUs. The third and fourth architectures, T-FERN and C-FERN, respectively, are models that add time selection and condition input extraction to FERN. We can observe from the Fig. \ref{fig:difArc} that while D-FERN improves the performance of FERN, both T-FERN and C-FERN can achieve a higher detection accuracy. The improvement is due to the fact that D-FERN only compares static and moving features for precise characteristic extraction, while T-FERN and C-FERN further extract and provide time variation features to better discriminate each case. With the effectiveness of total functionalities, our proposed TCD-FERN scheme can achieve the highest detection accuracy among all schemes. To elaborate a little further, in terms of the five testing locations, locations 1, 2, and 3 have higher accuracies compared to locations 4 and 5 as they are characterized by LoS path and reflective spaces, which provide more distinct features. However, location 4, which is an open space but close to the transmitter in \fig \ref{fig:scen1}, also demonstrates high accuracy. In contrast, location 5 is far from the transmitter, resulting in a weaker impact on the transmitted signal and the lowest accuracy among the five testing locations.


\subsection{Evaluation of Voting Scheme}
We evaluate the effectiveness of the online voting scheme to improve the detection accuracy of the TX room. We conduct the evaluation using the dataset of the second scenario collected in the conference room, which involves two transmission pairs for voting. The confusion matrices of transmission pair 1 and of transmission pair 2 are shown in Figs. \ref{fig:confusion1} and \ref{fig:confusion2}, respectively. The transmission pair 1 is with rooms 1 (TX) and 2 (RX), whereas transmission pair 2 is with rooms 1 (TX) and 3 (RX). From the confusion matrices, we can observe that when there are people in the room with RX, such as case 3, it is relatively challenging for the system to differentiate the human presence in the TX room. The merged probability of rooms 1 and 2 from the prediction of transmission pair 1 gives the accuracy of 96.78\% for room 1 and 98.45\% for room 2. Furthermore, the accuracies of 92.51\% for room 1 and 97.57\% for room 3 are obtained from the merged probability of transmission pair 2. Based on the accuracies obtained from the two transmission pairs, we can conclude that the TX room, which is room 1, consistently exhibits poorer performance compared to the RX rooms. However, the online voting scheme can effectively improve the accuracy of the TX room, especially in the case of transmission pair 2, where the accuracy of the TX room is increased from 80.46\% to 92.51\% with the help of the voting process. Thus, the evaluation confirms the efficiency of the online voting scheme in enhancing the detection accuracy of the TX room. The confusion matrix in \fig \ref{fig:confusionRoom1} shows the prediction accuracy for room 1 after the voting scheme is applied to both transmission pairs 1 and 2. Note that case 1 here indicates an empty room, whereas case 2 represents human presence. The final accuracy for room 1 is 97.35\%, which envisions the effectiveness of the voting approach.

\begin{table}[!t]
\footnotesize
\centering
\caption{Human Presence Scenarios}
\begin{tabular}{|c|c|c|c|c|}
\hline
\multicolumn{1}{|l|}{\multirow{2}{*}{}} & \multicolumn{4}{c|}{Data}                                                                                                                         \\ \cline{2-5} 
\multicolumn{1}{|l|}{}  & Class 1 & Class 2  & Class 3   & Class 4   \\ \hline
Scenario 1 & Empty  & Presence in TX    & -   & -  \\ \hline
Scenario 2  & Empty   & Both   & -  & -   \\ \hline
Scenario 3   & Empty  & Presence in TX & Presence in RX  & Both \\ \hline
\end{tabular}
\label{scenarios}
\end{table}

\begin{table*}[!t]
\scriptsize
\centering
\caption{Overall Performance Comparison}
\begin{tabular}{|l||cccc||cccc||cccc|}
\hline
 & \multicolumn{4}{c|}{Pair 1} & \multicolumn{4}{c|}{Pair 2} & \multicolumn{4}{c|}{Overall} \\ \hline
Scenario 1 & \multicolumn{1}{c|}{Accuracy} & \multicolumn{1}{c|}{Precision} & \multicolumn{1}{c|}{Recall} & F1-Score & \multicolumn{1}{c|}{Accuracy} & \multicolumn{1}{c|}{Precision} & \multicolumn{1}{c|}{Recall} & F1-Score & \multicolumn{1}{c|}{Accuracy} & \multicolumn{1}{c|}{Precision} & \multicolumn{1}{c|}{Recall} & F1-Score \\ \hline
R-TTWD & \multicolumn{1}{c|}{71.36} & \multicolumn{1}{c|}{67.48} & \multicolumn{1}{c|}{82.5} & 74.24 & \multicolumn{1}{c|}{68.125} & \multicolumn{1}{c|}{65.26} & \multicolumn{1}{c|}{77.5} & 70.86 & \multicolumn{1}{c|}{77.5} & \multicolumn{1}{c|}{78.2} & \multicolumn{1}{c|}{76.25} & 77.22 \\ \hline
HARNN (DT) & \multicolumn{1}{c|}{59.57} & \multicolumn{1}{c|}{59.77} & \multicolumn{1}{c|}{58.51} & 59.13 & \multicolumn{1}{c|}{61.16} & \multicolumn{1}{c|}{62.1} & \multicolumn{1}{c|}{57.29} & 59.6 & \multicolumn{1}{c|}{59.94} & \multicolumn{1}{c|}{69.09} & \multicolumn{1}{c|}{35.99} & 47.33 \\ \hline
TCD-FERN & \multicolumn{1}{c|}{\textbf{96.75}} & \multicolumn{1}{c|}{\textbf{96.06}} & \multicolumn{1}{c|}{\textbf{94.5}} & \textbf{96.77} & \multicolumn{1}{c|}{\textbf{92.75}} & \multicolumn{1}{c|}{89.77} & \multicolumn{1}{c|}{\textbf{96.5}} & \textbf{93.01} & \multicolumn{1}{c|}{\textbf{97.4}} & \multicolumn{1}{c|}{\textbf{95.19}} & \multicolumn{1}{c|}{\textbf{99}} & \textbf{97.06} \\ \hline\hline
Scenario 2 & \multicolumn{1}{c|}{Accuracy} & \multicolumn{1}{c|}{Precision} & \multicolumn{1}{c|}{Recall} & F1-Score & \multicolumn{1}{c|}{Accuracy} & \multicolumn{1}{c|}{Precision} & \multicolumn{1}{c|}{Recall} & F1-Score & \multicolumn{1}{c|}{Accuracy} & \multicolumn{1}{c|}{Precision} & \multicolumn{1}{c|}{Recall} & F1-Score \\ \hline
R-TTWD & \multicolumn{1}{c|}{87} & \multicolumn{1}{c|}{\textbf{98.25}} & \multicolumn{1}{c|}{84.17} & \textbf{90.66} & \multicolumn{1}{c|}{81} & \multicolumn{1}{c|}{84.57} & \multicolumn{1}{c|}{\textbf{91.33}} & 87.82 & \multicolumn{1}{c|}{79.75} & \multicolumn{1}{c|}{80.67} & \multicolumn{1}{c|}{78.25} & 79.44 \\ \hline
HARNN (DT) & \multicolumn{1}{c|}{80.49} & \multicolumn{1}{c|}{87.24} & \multicolumn{1}{c|}{86.67} & 86.95 & \multicolumn{1}{c|}{66.67} & \multicolumn{1}{c|}{79.67} & \multicolumn{1}{c|}{74.59} & 77.05 & \multicolumn{1}{c|}{68.34} & \multicolumn{1}{c|}{68.42} & \multicolumn{1}{c|}{68.14} & 68.28 \\ \hline
TCD-FERN & \multicolumn{1}{c|}{\textbf{99.75}} & \multicolumn{1}{c|}{\textbf{100}} & \multicolumn{1}{c|}{\textbf{99.67}} & \textbf{99.83} & \multicolumn{1}{c|}{\textbf{99.5}} & \multicolumn{1}{c|}{\textbf{99.34}} & \multicolumn{1}{c|}{\textbf{100}} & \textbf{99.67} & \multicolumn{1}{c|}{\textbf{98.5}} & \multicolumn{1}{c|}{\textbf{98.02}} & \multicolumn{1}{c|}{\textbf{99}} & \textbf{98.51} \\ \hline\hline
Scenario 3 & \multicolumn{1}{c|}{Accuracy} & \multicolumn{1}{c|}{Precision} & \multicolumn{1}{c|}{Recall} & F1-Score & \multicolumn{1}{c|}{Accuracy} & \multicolumn{1}{c|}{Precision} & \multicolumn{1}{c|}{Recall} & F1-Score & \multicolumn{1}{c|}{Accuracy} & \multicolumn{1}{c|}{Precision} & \multicolumn{1}{c|}{Recall} & F1-Score \\ \hline
R-TTWD & \multicolumn{1}{c|}{54.88} & \multicolumn{1}{c|}{53.58} & \multicolumn{1}{c|}{54.88} & 48.24 & \multicolumn{1}{c|}{43.38} & \multicolumn{1}{c|}{45.54} & \multicolumn{1}{c|}{43.38} & 42.09 & \multicolumn{1}{c|}{60.38} & \multicolumn{1}{c|}{57.4} & \multicolumn{1}{c|}{80.5} & 67.01 \\ \hline
HARNN (DT) & \multicolumn{1}{c|}{43.76} & \multicolumn{1}{c|}{43.8} & \multicolumn{1}{c|}{47.78} & 43.8 & \multicolumn{1}{c|}{32.7} & \multicolumn{1}{c|}{32.49} & \multicolumn{1}{c|}{32.7} & 32.49 & \multicolumn{1}{c|}{52.73} & \multicolumn{1}{c|}{52.9} & \multicolumn{1}{c|}{49.95} & 51.38 \\ \hline
TCD-FERN & \multicolumn{1}{c|}{\textbf{99.75}} & \multicolumn{1}{c|}{\textbf{100}} & \multicolumn{1}{c|}{\textbf{99.01}} & \textbf{99.5} & \multicolumn{1}{c|}{\textbf{99.5}} & \multicolumn{1}{c|}{\textbf{98}} & \multicolumn{1}{c|}{\textbf{100}} & \textbf{98.99} & \multicolumn{1}{c|}{\textbf{97.6}} & \multicolumn{1}{c|}{\textbf{96.35}} & \multicolumn{1}{c|}{\textbf{99}} & \textbf{97.66} \\ \hline
\end{tabular} \label{overperformance}
\end{table*}

\subsection{Benchmark Comparison}

To evaluate the efficiency of our proposed \netw for multi-room presence detection, we have compared our performance with existing algorithms R-TTWD \cite{23} and HARNN \cite{24}. R-TTWD proposes a robust through-the-wall detection of moving humans, where they collect the mean of first-order difference of eigenvectors as input feature and obtained a threshold by support vector machine (SVM) \cite{28}. HARNN proposes a human activity recognition network, where they utilized a decision tree with indications of variance and correlation coefficient for human presence detection. In this subsection, we compare our algorithm to the decision tree algorithm of HARNN for presence detection labeled as HARNN(DT). Although R-TTWD detects through-the-wall humans, both R-TTWD and HARNN apply presence detection to a single room. Hence, we compare the performance in three scenarios with different data grouping, which are shown in Table \ref{scenarios}. We use data from the conference room for all scenarios. For scenarios 1 and 2, we grouped data into two classes for classification, i.e., empty and human presence. The difference between these two scenarios is that we only detect human presence of the TX room in scenario 1, while we detect the presence of both TX and RX rooms in scenario 2. In scenario 3, there are four classes identical to the four cases in our system for multi-room classification.

Table \ref{overperformance} shows the performance comparison of the three algorithms based on transmission pairs 1 and 2 as well as overall averaged performance in terms of accuracy, precision, recall and F1-score. We can observe that all three algorithms in scenario 2 can achieve the highest accuracies compared to the other two scenarios, because high signal variance takes place between the cases of both rooms empty and both rooms with human presence. However, the classification in scenario 3 involves four classes of two rooms resulting in significant performance degradation on all metrics of R-TTWD and HARNN. Our proposed \netw can still provide better performance of above 95\% accuracy in both transmission pairs. Overall, the proposed TCD-FERN scheme outperforms the other existing benchmarks in terms of the highest overall accuracies of 97.4\%, 98.5\%, and 97.6\% in scenarios 1, 2 and 3, respectively. To elaborate a little further, we have analyzed the complexity of R-TTWD, HARNN, and TCD-FERN algorithms in Table \ref{TableComplexity}. We can observe that TCD-FERN possesses the highest complexity and the required memory size owing to its complex neural network architecture designs. Nevertheless, the proposed TCD-FERN can accomplish a substantially improvement of human presence detection accuracy of around $20\%$ to $40\%$ compared to other existing benchmarks. In conclusion, the merits of our proposed TCD-FERN scheme for multi-room human presence detection can therefore be revealed.

\begin{table}[!t]
\footnotesize
\centering
\caption{Complexity Analysis}
\begin{tabular}{|l|c|c|c|c|}
\hline
 & Parameter & \begin{tabular}[c]{@{}c@{}}Memory \\ Size\end{tabular} & \begin{tabular}[c]{@{}c@{}}Training \\ Time\end{tabular} & \begin{tabular}[c]{@{}c@{}}Testing \\ Time\end{tabular} \\ \hline
R-TTWD & 18 & 79.32KB & 0.36s & 0.05s \\ \hline
HARNN (DT) & 121,876 & 2.61KB & 0.15s & 0.005s \\ \hline
TCD-FERN & 506,988 & 6140KB & 28.6s & 4.8s \\ \hline
\end{tabular} \label{TableComplexity}
\end{table}

\section{Conclusion} \label{CHP_CON}

\subsection{Conclusion Remark}
In this paper, we present a novel CSI-based human presence detection system for multi-room environments, with the need of a single WiFi AP in each room. The \preproc data preprocess is designed to extract both moving and static human features. Our proposed \netw model addresses the challenges of insensitivity to human presence and distinguishes the cases between LoS and NLoS. It captures significant time features conditioned on the current human feature. To alleviate the problem of attenuated features at TX room due to partitioned walls and longer distance, we apply a probability-based voting process to enhance detection prediction accuracy. We evaluate different parameters and demonstrate the effectiveness of different architectural combinations in practical indoor scenarios with the deployment of commercial WiFi devices. Experimental results have shown that the voting scheme can achieve around $3\%$ improvement of human presence detection accuracy. It has also revealed the significant improvement of leveraging subcarrier fusion, dual-feature recurrent network, time selection and condition mechanisms. Compared to the existing works in open literature, our proposed TCD-FERN system can achieve above $97\%$ of human presence detection accuracy for multi-room scenarios with the adoption of fewer WiFi APs.

\subsection{Future Work}
In this work, performance is potentially limited by the hardware constraint, such as processing and transmission speed of APs as well as computing power of the edge server. The data dimension reduction will be conducted  in order to reduce complexity, which however may lead to insufficient feature information. This will degrade the detection performance in difficult classification tasks with massive labels. As future works, high-resolution CSI should be collected to distinguish complex activities, including pedestrian tracking, motion detection and vitality detection in multiple complicated rooms \cite{acm}. The next-generation WiFi-6/7 (802.11ax/be) AP device possessing higher bandwidth and faster processing speed is regarded as a potential of collecting fine-grained CSI and providing higher detection precision. However, the signal features may be indistinguishable owing to the similar activities in different rooms. With more complex activities, different statistical results as well as comprehensive information of other sensors can be considered as the detection input. Additionally, more advanced deep learning architecture can be leveraged, e.g., the transformer-based structures \cite{confT} with advanced attention capability can support regression and prediction of large models to harness the inter- and intra-dataset spatial and temporal correlations.

\footnotesize
\bibliographystyle{IEEEtran}
\bibliography{myReference}

\begin{thebibliography}{10}
\providecommand{\url}[1]{#1}
\csname url@samestyle\endcsname
\providecommand{\newblock}{\relax}
\providecommand{\bibinfo}[2]{#2}
\providecommand{\BIBentrySTDinterwordspacing}{\spaceskip=0pt\relax}
\providecommand{\BIBentryALTinterwordstretchfactor}{4}
\providecommand{\BIBentryALTinterwordspacing}{\spaceskip=\fontdimen2\font plus
\BIBentryALTinterwordstretchfactor\fontdimen3\font minus
  \fontdimen4\font\relax}
\providecommand{\BIBforeignlanguage}[2]{{%
\expandafter\ifx\csname l@#1\endcsname\relax
\typeout{** WARNING: IEEEtran.bst: No hyphenation pattern has been}%
\typeout{** loaded for the language `#1'. Using the pattern for}%
\typeout{** the default language instead.}%
\else
\language=\csname l@#1\endcsname
\fi
#2}}
\providecommand{\BIBdecl}{\relax}
\BIBdecl

\bibitem{ble_sense}
Y.~Yu, R.~Chen, L.~Chen, X.~Zheng, D.~Wu, W.~Li, and Y.~Wu, ``A novel 3-d
  indoor localization algorithm based on {BLE} and multiple sensors,''
  \emph{IEEE Internet of Things Journal}, vol.~8, no.~11, pp. 9359--9372, 2021.

\bibitem{phone_sen}
S.~Yang, J.~Liu, X.~Gong, G.~Huang, and Y.~Bai, ``A robust heading estimation
  solution for smartphone multisensor-integrated indoor positioning,''
  \emph{IEEE Internet of Things Journal}, vol.~8, no.~23, pp. 17\,186--17\,198,
  2021.

\bibitem{HanzoLifeTime}
H.~Yetgin, K.~T.~K. Cheung, M.~El-Hajjar, and L.~H. Hanzo, ``A survey of
  network lifetime maximization techniques in wireless sensor networks,''
  \emph{IEEE Communications Surveys Tutorials}, vol.~19, no.~2, pp. 828--854,
  2017.

\bibitem{DL_detect}
R.~Zhang, X.~Jing, S.~Wu, C.~Jiang, J.~Mu, and F.~R. Yu, ``Device-free wireless
  sensing for human detection: The deep learning perspective,'' \emph{IEEE
  Internet of Things Journal}, vol.~8, no.~4, pp. 2517--2539, 2021.

\bibitem{31}
X.~{Ma}, H.~{Wang}, B.~{Xue}, M.~{Zhou}, B.~{Ji}, and Y.~{Li}, ``Depth-based
  human fall detection via shape features and improved extreme learning
  machine,'' \emph{IEEE Journal of Biomedical and Health Informatics}, vol.~18,
  no.~6, pp. 1915--1922, 2014.

\bibitem{iotj_camera}
Y.~Zhao, J.~Xu, J.~Wu, J.~Hao, and H.~Qian, ``Enhancing camera-based multimodal
  indoor localization with device-free movement measurement using {WiFi},''
  \emph{IEEE Internet of Things Journal}, vol.~7, no.~2, pp. 1024--1038, 2020.

\bibitem{radar_fall}
B.~Wang, H.~Zhang, and Y.-X. Guo, ``Radar-based soft fall detection using
  pattern contour vector,'' \emph{IEEE Internet of Things Journal}, vol.~10,
  no.~3, pp. 2519--2527, 2023.

\bibitem{ourcsi1}
H.-C. Tsai, C.-J. Chiu, P.-H. Tseng, and K.-T. Feng, ``Refined
  autoencoder-based {CSI} hidden feature extraction for indoor spot
  localization,'' in \emph{Proc. IEEE Vehicular Technology Conference
  (VTC-Fall)}, 2018, pp. 1--5.

\bibitem{ourcsi2}
Y.-M. Huang, A.-H. Hsiao, C.-J. Chiu, K.-T. Feng, and P.-H. Tseng,
  ``Device-free multiple presence detection using {CSI} with machine learning
  methods,'' in \emph{Proc. IEEE Vehicular Technology Conference (VTC-Fall)},
  2019, pp. 1--5.

\bibitem{wifi_fall}
Y.~Wang, S.~Yang, F.~Li, Y.~Wu, and Y.~Wang, ``{FallViewer}: A fine-grained
  indoor fall detection system with ubiquitous wi-fi devices,'' \emph{IEEE
  Internet of Things Journal}, vol.~8, no.~15, pp. 12\,455--12\,466, 2021.

\bibitem{wifi_loc}
Y.~Yu, R.~Chen, L.~Chen, S.~Xu, W.~Li, Y.~Wu, and H.~Zhou, ``Precise 3-d indoor
  localization based on {Wi-Fi FTM} and built-in sensors,'' \emph{IEEE Internet
  of Things Journal}, vol.~7, no.~12, pp. 11\,753--11\,765, 2020.

\bibitem{wifi_sense2}
Y.~He, Y.~Chen, Y.~Hu, and B.~Zeng, ``{WiFi} vision: Sensing, recognition, and
  detection with commodity {MIMO-OFDM WiFi},'' \emph{IEEE Internet of Things
  Journal}, vol.~7, no.~9, pp. 8296--8317, 2020.

\bibitem{acm}
L.-H. Shen, K.-T. Feng, and L.~Hanzo, ``Five facets of {6G}: Research
  challenges and opportunities,'' \emph{ACM Computing Surveys}, vol.~55,
  no.~11, pp. 1--39, 2023.

\bibitem{1}
D.~Zhang and L.~M. Ni, ``Dynamic clustering for tracking multiple
  transceiver-free objects,'' in \emph{Proc. IEEE International Conference on
  Pervasive Computing and Communications (PerCom)}, 2009, pp. 1--8.

\bibitem{2}
P.~Bahl and V.~N. Padmanabhan, ``{RADAR}: An in-building {RF}-based user
  location and tracking system,'' in \emph{Proc. IEEE International Conference
  on Computer Communications (INFOCOM)}, vol.~2, 2000, pp. 775--784.

\bibitem{finger}
X.~Gong, J.~Liu, S.~Yang, F.~Gu, G.~Huang, and Y.~Bai, ``An enhanced indoor
  positioning solution using dynamic radio fingerprinting spatial context
  recognition,'' \emph{IEEE Internet of Things Journal}, vol.~10, no.~2, pp.
  1297--1309, 2023.

\bibitem{3}
Y.~Gu, F.~Ren, and J.~Li, ``{PAWS}: Passive human activity recognition based on
  {WiFi} ambient signals,'' \emph{IEEE Internet of Things Journal}, vol.~3,
  no.~5, pp. 796--805, 2015.

\bibitem{4}
H.~Abdelnasser, M.~Youssef, and K.~A. Harras, ``{WiGest}: A ubiquitous
  {WiFi}-based gesture recognition system,'' in \emph{Proc. IEEE International
  Conference on Computer Communications (INFOCOM)}, 2015, pp. 1472--1480.

\bibitem{5}
J.~Xiao, K.~Wu, Y.~Yi, and L.~M. Ni, ``{FIFS}: Fine-grained indoor
  fingerprinting system,'' in \emph{Proc. IEEE International Conference on
  Computer Communications and Networks (ICCCN)}, 2012, pp. 1--7.

\bibitem{6}
K.~Wu, J.~Xiao, Y.~Yi, M.~Gao, and L.~M. Ni, ``{FILA}: Fine-grained indoor
  localization,'' in \emph{Proc. IEEE International Conference on Computer
  Communications (INFOCOM)}, 2012, pp. 2210--2218.

\bibitem{7}
C.~Han, K.~Wu, Y.~Wang, and L.~M. Ni, ``{WiFall}: Device-free fall detection by
  wireless networks,'' in \emph{Proc. IEEE International Conference on Computer
  Communications (INFOCOM)}, 2014, pp. 271--279.

\bibitem{8}
Y.~Wang, K.~Wu, and L.~M. Ni, ``{WiFall}: Device-free fall detection by
  wireless networks,'' \emph{IEEE Transactions on Mobile Computing}, vol.~16,
  no.~2, pp. 581--594, 2017.

\bibitem{9}
W.~Wang, A.~X. Liu, M.~Shahzad, K.~Ling, and S.~Lu, ``Device-free human
  activity recognition using commercial {WiFi} devices,'' \emph{IEEE Journal on
  Selected Areas in Communications}, vol.~35, no.~5, pp. 1118--1131, 2017.

\bibitem{10}
J.~Yang, H.~Zou, H.~Jiang, and L.~Xie, ``Device-free occupant activity sensing
  using {WiFi}-enabled {IoT} devices for smart homes,'' \emph{IEEE Internet of
  Things Journal}, vol.~5, no.~5, pp. 3991--4002, 2018.

\bibitem{11}
S.~Tan and J.~Yang, ``{WiFinger}: Leveraging commodity {WiFi} for fine-grained
  finger gesture recognition,'' in \emph{Proc. ACM International Symposium on
  Mobile Ad Hoc Networking and Computing (MobiHoc)}, 2016, p. 201–210.

\bibitem{12}
J.~Yang, H.~Zou, Y.~Zhou, and L.~Xie, ``Learning gestures from {WiFi}: A
  siamese recurrent convolutional architecture,'' \emph{IEEE Internet of Things
  Journal}, vol.~6, no.~6, pp. 10\,763--10\,772, 2019.

\bibitem{13}
D.~Zhang, Y.~Hu, Y.~Chen, and B.~Zeng, ``{BreathTrack}: Tracking indoor human
  breath status via commodity {WiFi},'' \emph{IEEE Internet of Things Journal},
  vol.~6, no.~2, pp. 3899--3911, 2019.

\bibitem{14}
F.~Wang, F.~Zhang, C.~Wu, B.~Wang, and K.~J.~R. Liu, ``Respiration tracking for
  people counting and recognition,'' \emph{IEEE Internet of Things Journal},
  vol.~7, no.~6, pp. 5233--5245, 2020.

\bibitem{15}
H.~Zou, Y.~Zhou, J.~Yang, W.~Gu, L.~Xie, and C.~Spanos, ``{FreeCount}:
  Device-free crowd counting with commodity {WiFi},'' in \emph{Proc. IEEE
  Global Communications Conference (GLOBECOM)}, 2017, pp. 1--6.

\bibitem{16}
Z.~Tian, Y.~Li, M.~Zhou, and Z.~Li, ``{WiFi}-based adaptive indoor passive
  intrusion detection,'' in \emph{Proc. IEEE International Conference on
  Digital Signal Processing (DSP)}, 2018, pp. 1--5.

\bibitem{detect2}
J.~E. {Kim}, J.~H. {Choi}, and K.~T. {Kim}, ``Robust detection of presence of
  individuals in an indoor environment using {IR-UWB} radar,'' \emph{IEEE
  Access}, vol.~8, pp. 108\,133--108\,147, 2020.

\bibitem{detect2-1}
R.~{Zhou}, X.~{Lu}, P.~{Zhao}, and J.~{Chen}, ``Device-free presence detection
  and localization with {SVM} and {CSI} fingerprinting,'' \emph{IEEE Sensors
  Journal}, vol.~17, no.~23, pp. 7990--7999, 2017.

\bibitem{sense_bm1}
X.~Zeng, B.~Wang, C.~Wu, S.~D. Regani, and K.~J.~R. Liu, ``{WiCPD}: Wireless
  child presence detection system for smart cars,'' \emph{IEEE Internet of
  Things Journal}, vol.~9, no.~24, pp. 24\,866--24\,881, 2022.

\bibitem{mycolor}
C.-C. Hsieh, A.-H. Hsiao, C.-J. Chiu, and K.-T. Feng, ``{CSI} ratio with
  coloring-assisted learning for {NLoS} motionless human presence detection,''
  in \emph{Proc. IEEE Vehicular Technology Conference (VTC-Spring)}, 2022, pp.
  1--5.

\bibitem{ourcsichu}
F.-Y. Chu, C.-J. Chiu, A.-H. Hsiao, K.-T. Feng, and P.-H. Tseng, ``{WiFi}
  {CSI}-based device-free multi-room presence detection using conditional
  recurrent network,'' in \emph{Proc. IEEE Vehicular Technology Conference
  (VTC-Spring)}, 2021, pp. 1--5.

\bibitem{ourcsiKI}
K.-I. Lu, C.-J. Chiu, K.-T. Feng, and P.-H. Tseng, ``Device-free {CSI}-based
  wireless localization for high precision drone landing applications,'' in
  \emph{Proc. IEEE Vehicular Technology Conference (VTC-Fall)}, 2019, pp. 1--5.

\bibitem{17}
S.~Di~Domenico, M.~De~Sanctis, E.~Cianca, and M.~Ruggieri, ``{WiFi}-based
  through-the-wall presence detection of stationary and moving humans analyzing
  the doppler spectrum,'' \emph{IEEE Aerospace and Electronic Systems
  Magazine}, vol.~33, no. 5-6, pp. 14--19, 2018.

\bibitem{18}
J.~Wilson and N.~Patwari, ``See-through walls: Motion tracking using
  variance-based radio tomography networks,'' \emph{IEEE Transactions on Mobile
  Computing}, vol.~10, no.~5, pp. 612--621, 2011.

\bibitem{ttw1}
S.~D. Regani, B.~Wang, Y.~Hu, and K.~J.~R. Liu, ``{GWrite}: Enabling
  through-the-wall gesture writing recognition using {WiFi},'' \emph{IEEE
  Internet of Things Journal}, vol.~10, no.~7, pp. 5977--5991, 2023.

\bibitem{ttw2}
B.~Korany, H.~Cai, and Y.~Mostofi, ``Multiple people identification through
  walls using off-the-shelf {WiFi},'' \emph{IEEE Internet of Things Journal},
  vol.~8, no.~8, pp. 6963--6974, 2021.

\bibitem{29}
K.~Cho, B.~Van~Merrienboer, C.~Gulcehre, D.~Bahdanau, F.~Bougares, H.~Schwenk,
  and Y.~Bengio, ``Learning phrase representations using {RNN} encoder-decoder
  for statistical machine translation,'' \emph{arXiv preprint arXiv:1406.1078},
  2014.

\bibitem{30}
Y.~LeCun, Y.~Bengio, and G.~Hinton, ``Deep learning,'' \emph{Nature}, vol. 521,
  no. 7553, pp. 436--444, 2015.

\bibitem{27}
L.-H. Shen and K.-T. Feng, ``Joint beam and subband resource allocation with
  {QoS} requirement for millimeter wave {MIMO} systems,'' in \emph{Proc. IEEE
  Wireless Communications and Networking Conference (WCNC)}, 2017, pp. 1--6.

\bibitem{26}
Y.~G. Li and G.~L. Stuber, \emph{Orthogonal Frequency Division Multiplexing for
  Wireless Communications}.\hskip 1em plus 0.5em minus 0.4em\relax Springer
  Science \& Business Media, 2006.

\bibitem{25}
S.~Hochreiter and J.~Schmidhuber, ``Long short-term memory,'' \emph{Neural
  Computation}, vol.~9, no.~8, pp. 1735--1780, 1997.

\bibitem{cronos}
L.-H. Shen, C.-C. Hsieh, A.-H. Hsiao, and K.-T. Feng, ``{CRONOS}: Colorization
  and contrastive learning for device-free {NLoS} human presence detection
  using {Wi-Fi} {CSI},'' \emph{IEEE Internet of Things Journal}, pp. 1--1,
  2023.

\bibitem{21}
D.~Bahdanau, K.~Cho, and Y.~Bengio, ``Neural machine translation by jointly
  learning to align and translate,'' \emph{arXiv preprint arXiv:1409.0473},
  2014.

\bibitem{22}
M.~T. Luong, H.~Pham, and C.~D. Manning, ``Effective approaches to
  attention-based neural machine translation,'' \emph{arXiv preprint
  arXiv:1508.04025}, 2015.

\bibitem{conf}
S.-H. Chung, L.-H. Shen, and K.-T. Feng, ``Long-/short-term reinforcement
  learning for multi-aps channel allocation in {IEEE} 802.11ax {WLANs},'' in
  \emph{Proc. IEEE Wireless Communications and Networking Conference (WCNC)},
  2023, pp. 1--6.

\bibitem{23}
H.~Zhu, F.~Xiao, L.~Sun, R.~Wang, and P.~Yang, ``{R-TTWD}: Robust device-free
  through-the-wall detection of moving human with {WiFi},'' \emph{IEEE Journal
  on Selected Areas in Communications}, vol.~35, no.~5, pp. 1090--1103, 2017.

\bibitem{24}
J.~Ding and Y.~Wang, ``{WiFi CSI}-based human activity recognition using deep
  recurrent neural network,'' \emph{IEEE Access}, vol.~7, pp.
  174\,257--174\,269, 2019.

\bibitem{28}
B.~E. Boser, I.~M. Guyon, and V.~N. Vapnik, ``A training algorithm for optimal
  margin classifiers,'' in \emph{Proceedings the Fifth Annual Workshop on
  Computational Learning Theory}, 1992, pp. 144--152.

\bibitem{confT}
W.-Y. Chung, L.-H. Shen, K.-T. Feng, Y.-C. Lin, S.-C. Lin, and S.-F. Chang,
  ``{WiRiS}: Transformer for {RIS}-assisted device-free sensing for joint
  people counting and localization using {Wi-Fi CSI},'' in \emph{Proc. IEEE
  Annual International Symposium on Personal, Indoor and Mobile Radio
  Communications (PIMRC)}, 2023, pp. 1--6.

\end{thebibliography}

\end{document}